\documentclass[3p,11pt,sort&compress]{elsarticle}

\usepackage[titletoc,title]{appendix}
\usepackage{algorithm, algorithmic}
\usepackage{float}
\usepackage{enumerate}
\usepackage{amsmath, amsthm, amssymb, bm}
\usepackage{appendix}
\usepackage{graphicx}
\usepackage{wrapfig}
\usepackage{multirow}
\usepackage{caption}
\usepackage{subcaption}
\usepackage{hyperref}
\newcommand{\doiurl}[1]{DOI: \href{http://dx.doi.org/#1}{#1}}

\newcommand{\changespace}[1]{\renewcommand{\baselinestretch}{#1}\normalsize}

\begin{document}

\begin{frontmatter}

\title{Cooperative Group Optimization with Ants (CGO-AS): \\ 
Leverage Optimization with Mixed Individual and Social Learning}

\journal{Applied Soft Computing, 50: 223-234, 2017 {\em [\doiurl{10.1016/j.asoc.2016.11.018}]}}

\author[label1]{Xiao-Feng Xie}
\ead{xie@wiomax.com}
\author[label1,label2]{Zun-Jing Wang}
\ead{wang@wiomax.com}
 
\address[label1]{WIOMAX LLC, Pittsburgh, PA 15213}
\address[label2]{Department of Physics, Carnegie Mellon University, Pittsburgh, PA 15213}

\hypersetup{
  pdfinfo={
   Title={Cooperative Group Optimization with Ants (CGO-AS): Leverage Optimization with Mixed Individual and Social Learning},
   Author={Xiao-Feng Xie, Zun-Jing Wang}
  }
}

\begin{abstract}
\label{pap:abstract}
We present CGO-AS, a generalized Ant System (AS) implemented in the framework of Cooperative Group Optimization (CGO), to show the leveraged optimization with a mixed individual and social learning. Ant colony is a simple yet efficient natural system for understanding the effects of primary intelligence on optimization. However, existing AS algorithms are mostly focusing on their capability of using social heuristic cues while ignoring their individual learning. CGO can integrate the advantages of a cooperative group and a low-level algorithm portfolio design, and the agents of CGO can explore both individual and social search. In CGO-AS, each ant (agent) is added with an individual memory, and is implemented with a novel search strategy to use individual and social cues in a controlled proportion. The presented CGO-AS is therefore especially useful in exposing the power of the mixed individual and social learning for improving optimization. The optimization performance is tested with instances of the Traveling Salesman Problem (TSP). The results prove that a cooperative ant group using both individual and social learning obtains a better performance than the systems solely using either individual or social learning. The best performance is achieved under the condition when agents use individual memory as their primary information source, and simultaneously use social memory as their searching guidance. In comparison with existing AS systems, CGO-AS retains a faster learning speed toward those higher-quality solutions, especially in the later learning cycles. The leverage in optimization by CGO-AS is highly possible due to its inherent feature of adaptively maintaining the population diversity in the individual memory of agents, and of accelerating the learning process with accumulated knowledge in the social memory.

\end{abstract}


\end{frontmatter}

\section{Introduction}

Ants are extremely successful in evolution of intelligence. As shown by socio-biologists~\cite{Holldobler:1990p3697}, although each ant only has a minuscule brain, nontrivial primary components of intelligence in the collective context have been encoded in their navigational guidance systems using multiple simple information sources. For example, ants can communicate with others through indirect means of the pheromone trails that they deposited in their environment \cite{Deneubourg:1990p3701}.
The collective foraging behavior and strong exploitation capability in ant colonies has inspired the invention of various Ant System (AS) algorithms~\cite{Dorigo:1996p3275,Dorigo:1997p2698,Bonabeau:2000p3169}. Typical AS includes Ant Colony Optimization (ACS) \cite{Dorigo:1997p2698}, AS with ranking (AS$_\text{rank}$) \cite{bullnheimer1999improved}, and MAX-MIN ant system (MMAS) \cite{Stutzle:2000p2978}. In addition, there are diverse hybrid forms of AS with other optimization algorithms, such as PSO-ACO-3Opt \cite{mahi2015new}, ACO-ABC \cite{gunduz2015hierarchic}, and FOGS-ACO \cite{saenphon2014combining}. Among the existing algorithms of AS, the usage of pheromone trails in natural ants has attracted a broad research interest~\cite{Wehner:2003p3574,collett2012navigational,collett2014desert,cheng2014beginnings,bolek2015food}. 
Pheromone trails of ants have now been adopted as a paradigm by computational research communities to illustrate the emergence in self-organization~\cite{Deneubourg:1990p3701}. Though ignoring individual memory of ants, the algorithms of AS~\textemdash~mainly by the construction and usage of pheromone trails in natural ants~\textemdash~has displayed a remarkable optimizing capability, and has been applied with a great success to a large number of computationally hard problems, such as the traveling salesman problem (TSP) \cite{Stutzle:2000p2978,Reinelt:1991p2617,nallaperuma2015analyzing}, vehicle routing problems \cite{bullnheimer1999improved}, and mixed-variable problems \cite{liao2014ant}. 

Natural ants however build their intelligence with both social and individual learning. Socio-biologists have found that individual route memory of ants \cite{Giurfa:1999p3614,Wehner:2003p3574} plays a significant role in guiding the foraging of many natural ant species \cite{Harris:2005p3555,Macquart:2006p3720,sommer2008multiroute,bolek2015food,collett1998local}. A fairly great amount of natural ant species \cite{gruter2011decision,czaczkes2013ant} uses both collective pheromone trails and individual route memory in their navigational guidance systems, though it remains unclear how ants leverage their search in complex environments with such an integrated usage mixing the two memories which correspond respectively to their social and individual learning. 

In the present work, we aim to study the benefit using a mixed individual and social learning in AS systems. Since the existing AS algorithms have demonstrated the optimization power of social learning, what leverage can be gained in optimization by merely adding individual learning? For a better performance, how to distribute the social and individual learning if one mixes and uses them together as an integrated form of intelligence? In this sense, our computational experiments do not aspire to providing complete comparison with state of the art algorithms using various instances (e.g. across various algorithms and optimization problems). Rather, we attempt to understand the leveraging aspects in optimization from simply adding and mixing individual learning into its original solely-social-learning version of AS systems. This is because AS is a succinct but intrinsic model for understanding the role of learning. Given how technically involved an upgraded-learning-induced optimization improvement of an algorithm is, a question of considerable practical relevance is: How to effectively integrate individual learning into the system? The question is certainly not trivial. Realization of a mixed learning in an integrated form requires an algorithm being a more complex system which encompasses interactions among multiple memories (individual and social memories) and behaviors. A fundamental and effective support must be provided for maintaining the fast self-organized processes in such a mixed learning. 

In this study, we will approach the question with a specific framework, the Cooperative Group Optimization (CGO) framework~\cite{xie2014cooperative}. CGO was presented based on the nature-inspired paradigm of group problem solving \cite{Galef:1995p1128,Tomasello:1993p1330,Nemeth:1986p980,Dennis:1993p1298,Goncalo:2006p1071,Paulus:2000p1114,kavadias2009effects}, to explore high-quality solutions in the search landscape \cite{kauffman1987towards} of the problem to be solved. The agents of CGO not only exploit in a parallel way using diverse novel patterns \cite{Nemeth:1986p980} preserved in their individual memory \cite{Glenberg:1997p1390,Ericsson:1995p1364}, but also cooperate with their peers through the group memory \cite{Danchin:2004p1204,Dennis:1993p1298}. This means that each agent of CGO possesses a search capability through a mix of both individual and social learning \cite{Galef:1995p1128,Boyd2011,Tomasello:1993p1330}. Therefore, we use CGO to implement a generalized AS, called CGO-AS. The presented CGO-AS is especially suitable to realize the cooperative search using both individual route memory and pheromone trails, and to reproduce a navigational guidance system of natural ants. Moreover, CGO-AS provides an algorithmic implementation of AS systems in a generalized form of group problem solving, which is commonly used in advanced social groups including human groups. This is important, as the generalization of AS not only enables us to find the advanced strategies used by ants to strengthen their optimization, but also allows us to observe the potential nontrivial factors which might contribute to the primary form of group intelligence from the low-level cognitive colonies. We use the TSP \cite{Reinelt:1991p2617}, a well-known computationally hard problem, as the testing benchmark of performance for the comparison between CGO-AS and other existing AS systems as well as some recent published algorithms \cite{Stutzle:2000p2978,mahi2015new,saenphon2014combining,zhou2015discrete,ouaarab2014discrete}.

The remainder of the paper is organized as follows. In Section \ref{sec:ants}, we describe the studies of the navigational guidance components and systems used by natural ants in their foraging to digest the fundamental features of their learning behaviors which motivated AS systems and CGO-AS of this work. In Section \ref{sec:as}, we briefly introduce AS systems and one representative example, the MAX-MIN version.  In Section \ref{sec:cgo}, we outline CGO framework. In Section \ref{sec:cgo-as}, we present our CGO-AS imitating the natural ants with both social and individual learning, and describe how to use it for solving the TSP. In Section \ref{sec:results}, we present our experimental results showing the performance of the proposed CGO-AS approach, and discuss its features. Finally, we summarize our work.

\section{Real-World Ant Navigation}
\label{sec:ants}

Natural ants are important models for understanding the role of learning in evolution of intelligence and in the improvement of optimization technologies~\cite{Dorigo:1996p3275,Stutzle:2000p2978,Wehner:2003p3574}. Ant workers have miniature brains but often striking navigational performance and behavioral sophistication as individuals of socially complex cognitive colonies~\cite{Wehner:2003p3574}. Understanding the robust behavior of ants which solve complex tasks unveils parsimonious mechanism and architecture for general intelligence and optimization. Here we will first briefly review the usage of pheromone trails by ants in their foraging, which provided the foundation of existing AS systems~\cite{Dorigo:1996p3275,Stutzle:2000p2978}. Next, we will then describe the usage of individual memory by ants and their more advanced navigational guidance systems, which inspires the realization of CGO-AS in this work.

\subsection{Pheromone trails}

Many ant species can form and maintain pheromone trails \cite{Morgan:2009p3644,czaczkes2015trail}, even the volatile ones. The study of the fire ants has showed that pheromone trails provide feedback to ants for organizing the massive foraging at a colony level \cite{Holldobler:1990p3697}. Successful foragers deposit pheromone on their return trails to the nest, resulting in the effective trails strengthened since more workers add pheromone to it. on the contrary, the trail decays if its food runs out,  because foragers refrain from reinforcing the trail and the existing pheromone of the trail evaporates.  Pheromone trails provide ants a long-term memory of previously used trails, as well as a short-term attraction to recent rewarding trails \cite{Jackson:2006p3563}. 

Concerning pheromone trails of ants, a global adaptive process arises from the activities of many {\em agents} responding to local information in shared environments~\cite{Deneubourg:1990p3701}. The performance is achieved through the social learning of ants, with which ants have mutual interactions  via their pheromone trails. Computational models of ant systems \cite{Dorigo:1996p3275,Stutzle:2000p2978} have showed how ant workers could cooperate together through social learning via the pheromone trails, which exhibits an impressive optimization capability on some complex problems, such as the finding of short paths in the TSP. 

\subsection{Route memory}

Ants also navigate using vectors and landmark-based routes \cite{Giurfa:1999p3614,Wehner:2003p3574}, as shown in many species, for example, the wood ant ({\em Formica rufa}) \cite{Harris:2005p3555}, the tropical ant ({\em Gigantiops destructor}) \cite{Macquart:2006p3720}, the Australian desert ant ({\em Melophorus bagoti}) \cite{sommer2008multiroute}, and the North African desert ant ({\em Cataglyphis fortis})  \cite{bolek2015food,collett1998local}. In foraging, individual ants can obtain their routes by initial navigational strategies~\cite{collett1998local,Wehner:2003p3574}, can put their innate responses to landmarks \cite{Wehner:2003p3574}, and can also memorize early routes with their increasing experience \cite{Collett:2003p3648,Wolf:2008p3750}. These behaviors attribute to the individual learning ability of ants, which is fundamental for evolving the advanced forms of general intelligence~\cite{Glenberg:1997p1390}. 

A great deal of flexibility has been observed in the individual learning of ants on their route navigation~\cite{Holldobler:1990p3697}. Ants can steer by visual {\em landmarks} in their route navigation~\cite{Collett:2002p3603}. They instruct others when they recall particular steering cues~\cite{Wehner:2006p3740}. Ants can also learn path segments in terms of the associated {\em local vectors} that connect between landmarks \cite{collett1998local}. In addition, ants can memorize multiple routes \cite{sommer2008multiroute}, and can even steer the journeys that consist of separate path segments. Information combined from all experienced path segments may be used by ants as a memory network \cite{wystrach2013snapshots} to determine their familiar headings on given landmarks. Route memory often plays a significant role in guiding ants during their foraging activities.

\subsection{Navigational guidance system}

Ants integrate information from multiple sources in their navigational guidance systems \cite{Wehner:2003p3574,collett2012navigational,collett2014desert,cheng2014beginnings,bolek2015food} in order to efficiently search the paths between goals. For example, some ants \cite{gruter2011decision,czaczkes2013ant} use both pheromone trails and route memory in foraging. Notice that pheromone trails and route memory are respectively corresponding to the social and private individual information of ants that support their social and individual learning. Pheromone trails may cover more foraging paths by encoding the collective experiences of ants, but route memory is often more accurate in information than pheromone trails, even limited by the minuscule brains of ants. 

Natural ant colonies often use an integrated mixed learning in their foraging system, where route memory and pheromone trails combine together to a synergistic information cascades cooperatively providing an effective and efficient guidance over various foraging conditions. Route memories maintain a diversity of the high-quality information learned from individual experience of each ant, while pheromone trails provide a stability of the high-quality routes learned from all ants and over time~\cite{Collett:2002p3603}. The understanding on the real-world navigational guidance system of ants motivated us to present, implement and test CGO-AS system in this work.

\section{Ant Systems for the TSP}
\label{sec:as}

AS \cite{Dorigo:1996p3275} is a class of optimization algorithms inspired by the emergent search behavior using {\em pheromone trails} \cite{czaczkes2015trail} in natural ants~\cite{Goss:1989p3685}. Though different optimization problems \cite{Stutzle:2000p2978,Reinelt:1991p2617,bullnheimer1999improved,liao2014ant} have been solved with AS variants, the TSP is normally considered as a testing benchmark of ant navigation. 

The TSP \cite{Reinelt:1991p2617} can be described as a complete graph with $N$ nodes (or cities) and a cost matrix $D=(d_{ij})$, in which $d_{ij}$ is the length of an edge $(i, j)$ that connects between cities $i$ and $j$, where $i, j \in [1, N]$. The study here only concerns the symmetric TSP, which has $d_{ij}=d_{ji}$ for the edges. Each potential solution is a Hamiltonian tour $\bm{ \pi}=(\pi_{[1]}, \cdots, \pi_{[N]})$, which passes through each node once and only once, and its evaluation value $f(\bm{ \pi})$ is the total length of all edges in the tour. The optimization objective is to find a tour with the minimal evaluation value. 

In AS, there are a colony of $K$ artificial {\em ants}, where all ants search using a {\em pheromone matrix} $\varPsi=(\tau_{ij})$, in which $\tau_{ij}$ describes the pheromone trail from city $i$ to city $j$. The system runs in total $T$ iterations. At each iteration $t$, each ant builds its tour $\bm{ \pi}^{(t)}$ in an iterative way. As shown in Algorithm \ref{alg:antconstruct}, starting from a randomly selected city as the current city $i$, each ant chooses the next city $j$ to go with a probability biased by the pheromone trail $\tau_{ij}^the amount
{(t)}$ and by a locally available heuristic information $\eta_{ij}$ present on the connecting edge $(i, j)$, and continues this process till a tour is built. When an ant is at city $i$, the selection probability of the ant to city $j \in \mathcal{N}_{i}$ is described as \cite{Dorigo:1996p3275}: 

\begin{equation}\label{Eq:SelProb}
p_{ij}^{(t)}=\frac{[\tau_{ij}^{(t)}]^{\alpha}[\eta_{ij}]^{\beta}}{\sum_{l\in \mathcal{N}_{i}} [\tau_{il}^{(t)}]^{\alpha}[\eta_{il}]^{\beta}}~,
\end{equation}
where $\mathcal{N}_{i}$ is the candidate set of cities which the ant 
has not visited yet, and $\alpha$ and $\beta$ are two setting parameters which control
the relative importance of the pheromone trail and
heuristic information. By default, $\alpha=1$ and $\beta=2$. For the TSP, $\eta_{ij}$ is a function of the edge length, i.e., $\eta_{ij} = 1/d_{ij}$. Normally, the selection in Line 3 (see Algorithm \ref{alg:antconstruct}) is augmented with the candidate set of length 20 which contains the nearest neighbors \cite{Stutzle:2000p2978} to reduce the computational cost.

\begin{algorithm*} [htb]                     % enter the algorithm environment
\caption{Construct a tour by an ant using the pheromone matrix $\varPsi$}   % give the algorithm a caption
\label{alg:antconstruct}                           % and a label for \ref{} commands later
\begin{algorithmic}[1]                    % enter the algorithmic environment
\REQUIRE The pheromone matrix $\varPsi$ ~~\COMMENT{The cost matrix $D$ is a default input}
\STATE $\mathcal{N}=\{1, \cdots, N\}$; $i=\text{RND($\mathcal{N}$)}$; $\mathcal{N}_i=\mathcal{N}\backslash\{i\}$;  $\pi_{[1]}=i$~~\COMMENT{From a randomly selected city}
\FOR{$n$ = 2 to $N$} 
\STATE Select the next city $j\in \mathcal{N}_i$ with the probability $p_{ij}^{(t)}$, using Eq. \ref{Eq:SelProb} with $\varPsi$ and $D$
\STATE $\pi_{[n]}=j$; $\mathcal{N}_j=\mathcal{N}_i\backslash\{j\}$; $i=j$~~~~~~~~~~~~~~~~~~~~~~~~~~~~~~~~~~~~~~~~~~~~~\COMMENT{Move to the next city $j$} 
\ENDFOR
\RETURN $\bm{ \pi}=({\pi_{[1]}, \cdots, \pi_{[N]}})$ ~~~~~~~~~~~~~~~~~~~~~~~~~~~~~~~~~~~~~~~~~~~~~~\COMMENT{Return the new tour}
\end{algorithmic}
\end{algorithm*}

After all ants have constructed their tours, pheromone is updated on all edges as follow: 

\begin{equation}\label{Eq:UpdatePheromone}
\tau_{ij}^{(t+1)}=\rho\cdot\tau_{ij}^{(t)}+\sum_{k=1}^K \Delta \tau_{ij(k)}^{(t)}~,
\end{equation}
where the parameter $\rho \in [0, 1)$ is the trail persistence from evaporation, $\Delta \tau_{ij(k)}^{(t)}=1/f(\bm{ \pi}_{(k)}^{(t)})$ is the amount of pheromone which the ant $k$ puts on
the edge $(i, j)$ under the condition that the edge belongs to the tour $\bm{ \pi}_{(k)}^{(t)}$ done by the ant $k$ in the iteration $t$. By default, $\rho=0.5$.  

In Eq. \ref{Eq:UpdatePheromone}, evaporation mechanism enables the system to “forget” unuseful edges over time, and a greater amount of pheromone is allocated to the shorter tours. In Eq. \ref{Eq:SelProb}, the selection probability is achieved from a combination of the global heuristic cue (of tour length) from the pheromone trail $\tau_{ij}(t)$ and the local heuristic cue (of edge length) from the heuristic information $\eta_{ij}$. Edges which are contained in the shorter tours will receive more pheromone and thus will be chosen by ants with higher probabilities in future iterations. 

The ants in AS do not possess long-term individual memory. Rather, pheromone matrix $\varPsi$ plays the role of a long-term social memory distributed on the edges of the graph, which is iteratively modified by ants to reflect their experience accumulated in solving the problem. This allows an indirect form of learning called {\em stigmergy} \cite{Dorigo:1997p2698}. 

\subsection{MAX-MIN Ant System (MMAS)}

MAX-MIN ant system (MMAS) \cite{Stutzle:2000p2978} is one of the best performing ant systems. It has been specifically developed for achieving a better performance by the combination between an improved exploitation of the best solutions found in search and an effective mechanism which leads to the choice probabilities avoiding early search stagnation. 

MMAS differs in two key aspects from the original AS. First, in order to impose strong exploitation on the best solutions found in search, after each iteration, only one ant deposits pheromone on the best solution, either in the current iteration ({\em iteration-best} solution $\bm{ \pi}^{(t)}_{\text{ib}}$) or from the beginning ({\em best-so-far} solution $\bm{ \pi}^{(t)}_{\text{gb}}$). Second, in order to prevent a search from stagnation, the range of possible pheromone trails is limited within an interval $[\tau_{\text{min}} , \tau_{\text{max}}]$. The pheromone trails are initialized to be $\tau_{\text{max}}$ to achieve a higher exploration of solutions at the beginning of the search.

The values of $\tau_{\text{max}}$ and $\tau_{\text{min}}$ are respectively defined as \cite{Stutzle:2000p2978}:
\begin{equation}
\tau_{\text{max}}^{(t)}=\frac{1}{(1-\rho)f(\bm{ \pi}^{(t)}_{\text{gb}})}~,
\end{equation}
\begin{equation} 
\tau_{\text{min}}^{(t)}=\frac{\tau_{\text{max}}(p_{\text{best}}^{-1/N}-1)}{n/2-1}~,
\end{equation}
where $f(\bm{ \pi}^{(t)}_{\text{gb}})$ is the evaluation value of the best-so-far solution $\bm{ \pi}^{(t)}_{\text{gb}}$ at the iteration $t$, $\rho$ is the trail
persistence,  $p_{\text{best}} \in [(N/2)^{-N}, 1]$ is a setting parameter. If $p_{\text{best}}=1$, then $\tau_{\text{min}}=0$. The value of $\tau_{\text{min}}$ increases as $p_{\text{best}}$ decreases. By default, $p_{\text{best}}=0.05$.

\section{Cooperative Group Optimization (CGO)}
\label{sec:cgo}

CGO is an optimization framework based on the nature-inspired paradigm of group problem solving~\cite{xie2014cooperative}. With CGO, optimization algorithms can be represented in a script form using embedded search heuristics (ESHs) with the support of memory protocol specification (MP-SPEC) for the group of agents, using a toolbox of knowledge components. CGO has been used to describe some existing algorithms and realize their hybrids on solving numerical optimization problems.

\begin{figure*} [htb]
\centering \includegraphics[width=2.6in]{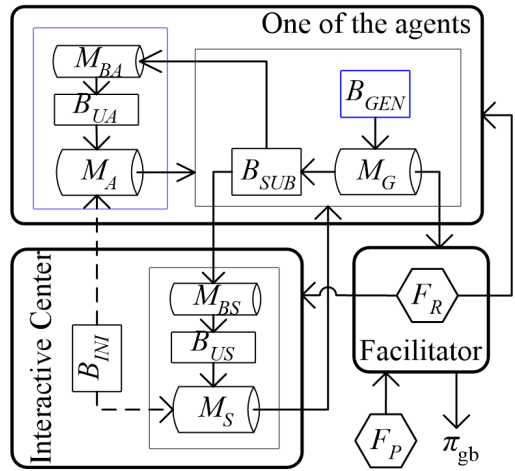} \caption{Simplified CGO Framework for CGO-AS.}
\label{Fig:cgos-as-01}
\end{figure*}

Figure \ref{Fig:cgos-as-01} gives a simplified version of CGO used in this paper for CGO-AS. The framework consists of a group of $N$ {\em agents}, an {\em interactive center} (IC), and a {\em facilitator}. It runs iteratively as a Markov chain in total $T$ learning cycles. 

The components in Figure \ref{Fig:cgos-as-01} are defined with a list of acronyms as:\\  

$F_P$: The optimization problem to be solved

$F_R$: The internal representation of $F_P$ in the form of a {\em search landscape} %$<S, R_{M}, AUX>$

$\bm{ \pi}_\text{gb}$: the \emph{best-so-far} solution found for $F_P$

$M_A$: The {\em private memory} of each agent, which can be accessed and is updated by the agent

$M_{BA}$: A buffer for storing the chunks generated by the agent for updating $M_A$ in each cycle

$B_{UA}$: The $M_A${\em-updating} behavior to update $M_A$ using the information (chunks) in $M_{BA}$

$M_G$: The {\em generative buffer} of each agent, which store newly generated information in each cycle

$M_S$: The {\em social memory} in IC, which can be accessed by the agents and is maintained by IC

$M_{BS}$: A buffer for collecting information (chunks) generated by the agents for updating $M_S$

$B_{US}$: The $M_S${\em-updating} behavior to update $M_S$ using the chunks in $M_{BS}$

$B_{INI}$: The {\em initializing} behavior to initialize information in $M_A$ of the agents and $M_S$ of the IC

$B_{GEN}$: The {\em generating} behavior of each agent, which generates new chunks into $M_G$ of the agent, using the mixed information (chunks) from $M_A$ of the agent and $M_S$ in IC, in each cycle

$B_{SUB}$: The {\em submitting} behavior of each agent, which submits the chunks in $M_G$ to $M_{BA}$ of the agent and to $M_{BS}$ in IC, in each learning cycle

\subsection{Facilitator}

The facilitator manages basic interfaces for the optimization problem $F_P$ to be solved. An essential landscape for the problem can be represented as a tuple $\langle S, R_{M}, AUX\rangle$. $S$ is the {\em problem space}, in which each  {\em state} $\bm{ \pi}\in S $ is a potential solution. $\forall \bm{ \pi}_{(a)} ,\bm{ \pi}_{(b)} \in S $, the \emph{quality-measuring} rule ($R_{M}$) measures the \emph{quality} difference between them. If the quality of $\bm{ \pi}_{(a)}$ is better than that of $\bm{ \pi}_{(b)} $, then $R_{M}$($\bm{ \pi}_{(a)} $,$\bm{ \pi}_{(b)}$) returns TRUE, otherwise it returns FALSE. $AUX$ contains all auxiliary components associated with the structural information of the problem.

The facilitator has two basic roles. The first role is to update the \emph{best-so-far} solution $\bm{ \pi}_\text{gb}^{(t)}$, by storing the better-quality state between $\bm{ \pi}_\text{gb}$ and each newly generated state using the $R_{M}$ rule. The second role is to provide a {\em search landscape}, i.e., $F_R=<S, R_{M}, AUX>$, which includes all heuristic cues of the problem that is useful for reaching the high-quality states. 

\subsection{Agents and IC}

Searching on the search landscape is performed by agents with the support from IC. The general solving capability arises from the interplay between {\em memory} ($M$) and {\em behavior} ($B$) \cite{Anderson:2005p1258} owned by these entities. The actual implementation is flexibly defined by a symbolic {\em script} over a {\em toolbox} of knowledge element instances.

\subsubsection{Memory and Behavior}

Each memory \cite{Glenberg:1997p1390} contains a list of cells storing basic declarative knowledge elements, called {\em chunks} ($CH$)\cite{Anderson:2005p1258}, which are associated with the information in the search landscape of $F_R$. During a runtime, each memory can be only updated by its owner. Each behavior, performed by its owner, applies {\em rule}(s) ($R$) to interact with some chunks in memory during a learning process.

There are three essential memories in CGO framework. IC maintains a social memory ($M_S$). Each agent possesses a private memory ($M_A$) and a generative buffer ($M_G$). $M_A$ and $M_G$ can be accessed only by its owner, while $M_S$ can be accessed by all agents. Both $M_S$ and $M_A$ are long-term memory (LTM) to hold chunks over learning cycles, while $M_G$ is a buffer for new chunks and will be cleared at the end of each learning cycle. 

IC holds two basic behaviors. The {\em initializing} behavior ($B_{INI}$) is used to initialize chunks in $M_A$ of the agents and $M_S$ of the IC. IC also holds a buffer $M_{BS}$ for collecting chunks from the agents. The $M_S${\em-updating} behavior ($B_{US}$) updates $M_S$ by using the chunks in the buffer $M_{BS}$. 

The search process for solving $F_R$ is performed by agents. Each agent has the following basic behaviors:
(a) The {\em generating} behavior ($B_{GEN}$) can generate new chunks into a generative buffer ($M_G$), using the chunks in both its $M_A$ and $M_S$ in IC; (b) the {\em submitting} behavior ($B_{SUB}$) is used to submit chunks in $M_G$ to $M_{BA}$ of the agent and to $M_{BS}$ in IC; (c) The $M_A${\em-updating} behavior ($B_{UA}$) is applied to update $M_A$ using elements obtained in $M_{BA}$; and (d) The state(s) in $M_G$ are extracted and exported to facilitator as the candidates for potential solution(s). In each learning cycle, $B_{GEN}$ is performed at first, and the newly generated elements in $M_G$ are processed afterwards by agents with the other three behaviors. 

\subsubsection{Script Representation}

For agents and IC, the essential search loop is driven by {\em embedded search heuristic} (ESH) with the support of {\em memory protocol specification} (SPEC-MP). SPEC-MP is used to define how chunks will be initialized and updated in $M_{A}$ of each agent and $M_{S}$ of IC, given that chunks are newly generated in $M_{G}$ of the agents, and we need to maintain the consistency of the interactions between memory and behavior. By defining the part to generate new chunks in $M_{G}$ using the chunks in $M_{A}$ of each agent and $M_{S}$ of IC, each ESH is able to close the search loop. Therefore, each ESH can be seen as a stand-alone algorithm instance with a solid low-level support of SPEC-MP in the framework of CGO. 

SPEC-MP contains a table of \emph{memory protocol rows}, where each row contains five elements, i.e., $\langle ID_{M},CH_{M},R_{IE},R_{UE},CH_{U} \rangle$.
$ID_{M}\in \{M_{A}, M_{S}\}$ refers to a long-term memory. $CH_{M}$ is a unique chunk in the memory $ID_{M}$. Each row thus refers to the chunk $CH_{M}$ in the memory $ID_{M}$. $CH_{U}$ is a chunk in $M_{A}$ and $M_{G}$. Each {\em elemental initializing} ($R_{IE}$) rule is used to output one chunk $CH_{(I)}$ for initializing $CH_{M}$. Each {\em elemental updating} ($R_{UE}$) rule updates the chunk $CH_{(M)}$ by taking two inputs $(CH_{(M)}, CH_{(U)})$. 
SPEC-MP can be split into two subtables, SPEC-MP$_A$ and SPEC-MP$_S$, where their $ID_M$ are respectively $M_A$ and $M_S$. 
Notice the fact that there are multiple agents but only one IC. 
Corresponding to $ID_{M}$ in $M_{A}$ and $M_{S}$, the types for $CH_{(I)}$ of $R_{IE}$ are respectively $\$CH_{M}$ and $CH_{M}$, the types for $CH_{(U)}$ of $R_{UE}$ are respectively $CH_{M}$ and $\$CH_{M}$, and the types for $CH_{(M)}$ of $R_{UE}$ are both $CH_{M}$. Here $\$CH$ means a set of chucks of the type $CH$. 

Each row in SPEC-MP defines an updatable relation from $CH_{U}$ to $CH_{M}$. The validity of all updatable relations can be easily checked with an {\em updatable graph} \cite{xie2014cooperative} which uses the chunks in $\{M_{A}, M_{S}, M_{G}\}$ as its nodes, and use updatable relations as its directed arcs. Since the chunks in $M_{G}$ are generated in learning cycles, each chunk in $M_{A}$ and $M_{S}$ can be updated only if this chunk has a directed path originating from a chunk in $M_{G}$. 

Each ESH is defined as $\langle R_{GE},E_{IG},CH_{OG} \rangle$. $R_{GE}$ is an {\em elemental generating} rule. $E_{IG}$ is an ordered list of chunks, where each chuck belongs to $M_A$ or $M_S$. $CH_{OG}$ is a chunk in $M_G$. The $R_{GE}$ rule takes $E_{IG}$ as its input, and outputs $CH_{OG}$ to $M_G$.

The chunks in $M_A$, $M_S$, and $M_G$ are of some primary chunk interfaces, and there are three primary rule interfaces, i.e., $R_{IE}$, $R_{UE}$, and $R_{GE}$ rules. Knowledge components of these primary chunk and rule interfaces could be implemented in the {\em toolbox} of CGO, and each instance could be called symbolically using its identifier and setting parameters. 

Notice that, different optimization algorithms, from simple to complex, can be easily implemented at the symbolic layer using SPEC-MP and ESHs to call the instances in CGO toolbox. 

\subsection{Execution Process}

Algorithm \ref{alg:cgoframework} gives the essential process for executing a single ESH with the support of SPEC-MP in the framework of CGO, where the working module (entity), the required inputs, and the outputs or updated modules are provided. In Line 1, $F_P$ is formulated into the form of a static search landscape $F_R=\langle S, R_{M}, AUX \rangle$. In Lines 2 and 3, all long-term memories used by the agents and IC are initialized by using $B_{INI}$, based on SPEC-MP.
After the initialization, the framework of CGO runs in the form of iterative learning cycles, where each learning cycle $t\in [1, T]$ is executed between Lines 4--10. In lines 5--8, each agent $k\in[1, K]$ is executed. In Line 5, given the ESH, $B_{GEN}$ is executed to generate the output chunk $CH_{OG} \in M_{G(k)}$, using the chunks in $M_{A(k)}$ and $M_{S}$. In Line 6, $B_{SUB(k)}$ is applied to submit $CH_{U}\in M_{A(k)}\cup M_{G(k)}$ into the buffer $M_{BA(k)}$ of the agent and into the buffer $M_{BS}$ to form a chunk set $\$CH_{U}$, based on SPEC-MP. In Line 7, $B_{UA(k)}$ is executed to update $CH_{M} \in M_{A(k)}$ using the chunks stored in the buffer $M_{BA(k)}$, based on SPEC-MP. In Line 8, the chunk contained in $M_{G(k)}$ is processed by facilitator to obtain the best-so-far solution $\bm{ \pi}_\text{gb}$. In Line 10, $B_{US}$ is executed to update $M_{S}$ using the chunks collected in the buffer $M_{BS}$, based on SPEC-MP. Finally, $\bm{ \pi}_\text{gb}^{(T)}$ is returned and the framework is terminated. 

\begin{algorithm*} [htb]                     % enter the algorithm environment
\caption{The execution process of the essential CGO framework}   % give the algorithm a caption
\label{alg:cgoframework}                           % and a label for \ref{} commands later
\begin{algorithmic}[1]                    % enter the algorithmic environment
\REQUIRE The optimization problem $F_P$~\COMMENT{Other inputs: SPEC-MP, ESH, and a toolbox of chunk and rule instances}
\STATE Facilitator: $F_P \rightarrow F_R=\langle S, R_{M}, AUX \rangle$~\COMMENT{Search landscape used by the behaviors of all entities}
\STATE $B_{INI}$: $\text{SPEC-MP} \rightarrow \langle \{M_{A(k)}|k\in[1, K]\}, M_{S}\rangle$ ~\COMMENT{Initialize memories at $t$=0}
\FOR{$t$ = 1 to $T$} 
\FOR{$k$ = 1 to $K$} 
\STATE $B_{GEN(k)}$: $\langle \text{ESH}, M_{A(k)}, M_{S} \rangle \rightarrow CH_{OG} \in M_{G(k)}$~\COMMENT{Agent $k$: Generate a new chunk}
\STATE $B_{SUB(k)}$: $\langle\text{SPEC-MP}, M_{A(k)}, CH_{OG} \rangle \rightarrow \langle M_{BA(k)}, M_{BS} \rangle$~\COMMENT{Agent $k$: Submit chunk(s) to  $M_{BA(k)}$ \& $M_{BS}$}
\STATE $B_{UA(k)}$: $\langle \text{SPEC-MP}, M_{A(k)}^{(t)},  M_{BA(k)} \rangle \rightarrow$ $M_{A(k)}^{(t+1)}$~\COMMENT{Agent $k$: Update $M_{A}$}
\STATE Facilitator: $\bm{ \pi}_\text{gb}^{(t)} \leftarrow  CH_{OG}$ {\bf if} $R_M ( CH_{OG}, \bm{ \pi}_\text{gb}^{(t)})\equiv$ TRUE~\COMMENT{Update $\bm{ \pi}_\text{gb}^{(t)}$}
\ENDFOR
\STATE $B_{US}$: $\langle \text{SPEC-MP}, M_{S}^{(t)}, M_{BS} \rangle \rightarrow M_{S}^{(t+1)}$\COMMENT{IC: Update $M_S$ (All buffer memories are cleared at the end of cycle)}
\ENDFOR
\RETURN $\bm{ \pi}_\text{gb}^{(T)}$ ~\COMMENT{Return the best-so-far solution}
\end{algorithmic}
\end{algorithm*}

Further details on the execution of the major behaviors, i.e., $B_{INI}$, $B_{GEN}$, $B_{SUB}$, $B_{UA}$ and $B_{US}$, are respectively provided in Algorithms \ref{alg:B_INI} - \ref{alg:B_US}. Notice that only $B_{GEN}$ is driven by ESH = $\langle R_{GE},E_{IG},CH_{OG} \rangle$, which uses the $R_{GE}$ rule to generate a new chunk $CH_{OG} \in M_G$, see Algorithms \ref{alg:B_GEN}.  
All the other behaviors are driven by SPEC-MP, which maintain $M_A$ of the agents and $M_S$ of the IC, see Algorithms \ref{alg:B_INI}, \ref{alg:B_SUB}--\ref{alg:B_US}. Each row of SPEC-MP contains $\langle ID_{M},CH_{M},R_{IE},R_{UE},CH_{U} \rangle$. %, as shown in Algorithms \ref{alg:B_INI}, \ref{alg:B_SUB} - \ref{alg:B_US}. 
For convenience, SPEC-MP is split into two subtables, SPEC-MP$_A$ and SPEC-MP$_S$, where their $ID_M$ are respectively $M_A$ and $M_S$. 
The elements in the $m$th row of SPEC-MP$_A$ and the $n$th row of SPEC-MP$_S$ are respectively indexed by subscripts $(A, m)$ and $(S,n)$. The chunks $M_{A(m)}$ and $M_{S(n)}$, i.e., the $m$th chunk in $M_{A}$ and the $n$th chunk in $M_S$, are respectively defined by $CH_{M(A, m)}$ and $CH_{M(S, n)}$. For each row of SPEC-MP, the $B_{INI}$ behavior uses the $R_{IE}$ rule to initialize the chunk $CH_{M}$, see Algorithm \ref{alg:B_INI}. For each new chunk in $M_G$, $B_{SUB}$ allocates each $CH_{U}$ into the buffers, see Algorithm \ref{alg:B_SUB}. Then $B_{UA}$ of each agent and $B_{US}$ of the IC use each $R_{UE}$ rule to update each chunk in the memories, see Algorithms \ref{alg:B_UA}--\ref{alg:B_US}. 

\begin{algorithm*} [p]                     % enter the algorithm environment
\caption{Execution of the {\em initializing} behavior $B_{INI}$}   % give the algorithm a caption
\label{alg:B_INI}                           % and a label for \ref{} commands later
\begin{algorithmic}[1]                    % enter the algorithmic environment
\REQUIRE SPEC-MP=$\langle$SPEC-MP$_A$, SPEC-MP$_S \rangle$~\COMMENT{Each row of SPEC-MP is $\langle ID_{M},CH_{M},R_{IE},R_{UE},CH_{U} \rangle$}
\FOR[For each row of SPEC-MP$_A$, the subtable of SPEC-MP with $ID_M$=$M_A$]{$m$ = 1 to $|$SPEC-MP$_A|$} 
\STATE $\{CH_{M(k)} |k\in[1, K]\}$ = $R_{IE(A,m)}()$~\COMMENT{Generate a chunk set of the type $\$CH_M$ with $K$ elements}
\STATE {\bf for}~~{$k$ = 1 to $K$}~~{\bf do}~~$CH_{M(k)}\rightarrow M_{A(k,m)}$~~{\bf end for} ~\COMMENT{Initialize $M_A$ of all agents using the chunks in the set}
\ENDFOR
\FOR[For each row of SPEC-MP$_S$, the subtable of SPEC-MP with $ID_M$=$M_S$]{$n$ = 1 to $|$SPEC-MP$_S|$} 
\STATE $CH_{M}$ = $R_{IE(S,n)}()$;~~$CH_{M}\rightarrow M_{S(n)}$ ~\COMMENT{Generate a chunk of the type $CH_M$; and initialize $M_S$ of IC using the chunk}
\ENDFOR
\RETURN $\{M_{A(k)}|k\in[1, K]\}, M_{S}$~\COMMENT{The $M_{(A)}$ of all agents and $M_S$ in IC are initialized}
\end{algorithmic}
\end{algorithm*}

\begin{algorithm*} [p]                     % enter the algorithm environment
\caption{Execution of the {\em generating} behavior $B_{GEN}$ of the agent $k$}   % give the algorithm a caption
\label{alg:B_GEN}                           % and a label for \ref{} commands later
\begin{algorithmic}[1]                    % enter the algorithmic environment
\REQUIRE $\langle \text{ESH}, M_{A(k)}, M_{S} \rangle$~\COMMENT{ESH=$\langle R_{GE},E_{IG},CH_{OG} \rangle$}
\STATE $\langle M_{A(k)}, M_{S}\rangle \rightarrow E_{IG}$~\COMMENT{Collect the ordered chunk list $E_{IG}$ using the chunks from $M_{A(k)} \cup M_{S}$}
\STATE $CH_{OG} =R_{GE}(E_{IG})$~\COMMENT{Generate new chunk $CH_{OG}$ by the instance of $G_{GE}$ rule, using $E_{IG}$ as the input} 
\STATE $CH_{OG}\rightarrow M_{G(k)}$~\COMMENT{Store the newly generated chunk $CH_{OG}$ into $M_{G(k)}$}
\end{algorithmic}
\end{algorithm*}

\begin{algorithm*} [p]                     % enter the algorithm environment
\caption{Execution of the {\em submitting} behavior $B_{SUB}$ of the agent $k$}   % give the algorithm a caption
\label{alg:B_SUB}                           % and a label for \ref{} commands later
\begin{algorithmic}[1]                    % enter the algorithmic environment
\REQUIRE $\langle\text{SPEC-MP}, M_{A(k)}, CH_{OG} \rangle$~\COMMENT{Each row of SPEC-MP is $\langle ID_{M},CH_{M},R_{IE},R_{UE},CH_{U} \rangle$}
\FOR[For each row of SPEC-MP$_A$, the subtable of SPEC-MP with $ID_M$=$M_A$]{$m$ = 1 to $|$SPEC-MP$_A|$} 
\STATE $\langle M_{A{(k)}}, CH_{OG} \rangle \rightarrow CH_{U(A,m)}$~\COMMENT{Retrieve the chunk $CH_{U(A,m)}$ from $M_{A{(k)}}\cup CH_{OG}$} 
\STATE $M_{BA(k,m)} = CH_{U(A,m)}$~\COMMENT{Put $CH_{U(A,m)}$ as the $m$th chunk of $M_{BA(k)}$} 
\ENDFOR
\FOR[For each row of SPEC-MP$_S$, the subtable of SPEC-MP with $ID_M$=$M_S$]{$n$ = 1 to $|$SPEC-MP$_S|$} 
\STATE $\langle M_{A{(k)}}, CH_{OG} \rangle \rightarrow CH_{U(S,n)}$~\COMMENT{Retrieve the chunk $CH_{U(S,n)}$ from $M_{A{(k)}}\cup CH_{OG}$} 
\STATE $M_{BS(n)} = CH_{U(S,n)} \cup M_{BS(n)}$~\COMMENT{Collect $CH_{U(S,n)}$ into the $n$th cell of $M_{BS}$} 
\ENDFOR
\RETURN $M_{BA(k)}, M_{BS}$~\COMMENT{$M_{BA(k)}$ is filled by the agent $k$, whereas $M_{BS}$ is filled by all the agents}
\end{algorithmic}
\end{algorithm*}

\begin{algorithm*} [p]                     % enter the algorithm environment
\caption{Execution of the $M_A${\em-updating} behavior $B_{UA}$ of the agent $k$}   % give the algorithm a caption
\label{alg:B_UA}                           % and a label for \ref{} commands later
\begin{algorithmic}[1]                    % enter the algorithmic environment
\REQUIRE $\langle \text{SPEC-MP}_A, M_{A(k)}^{(t)},  M_{BA(k)} \rangle$ ~\COMMENT{Each row of SPEC-MP is $\langle ID_{M},CH_{M},R_{IE},R_{UE},CH_{U} \rangle$}
\FOR[For each row of SPEC-MP$_A$, the subtable of SPEC-MP with $ID_M$=$M_A$]{$m$ = 1 to $|$SPEC-MP$_A|$} 
\STATE $M_{A(k,m)}^{(t+1)} = R_{UE(A,m)}(M_{A(k,m)}^{(t)}, M_{BA(k,m)})$~\COMMENT{Update the $m$th chunk in $M_{A(k)}$ using the $m$th chunk in $M_{BA(k)}$}
\ENDFOR
\RETURN $M_{A(k)}^{(t+1)}$~\COMMENT{$M_{A(k)}^{(t)}$ is updated into $M_{A(k)}^{(t+1)}$}
\end{algorithmic}
\end{algorithm*}

\begin{algorithm*} [p]                     % enter the algorithm environment
\caption{Execution of the $M_S${\em-updating} behavior $B_{US}$ of IC}   % give the algorithm a caption
\label{alg:B_US}                           % and a label for \ref{} commands later
\begin{algorithmic}[1]                    % enter the algorithmic environment
\REQUIRE $\langle \text{SPEC-MP}_S, M_{S}^{(t)}, M_{BS} \rangle$~\COMMENT{Each row of SPEC-MP is $\langle ID_{M},CH_{M},R_{IE},R_{UE},CH_{U} \rangle$}
\FOR[For each row of SPEC-MP$_S$, the subtable of SPEC-MP with $ID_M$=$M_S$]{$n$ = 1 to $|$SPEC-MP$_S|$} 
\STATE $M_{S(n)}^{(t+1)} = R_{UE(S,n)}(M_{S(n)}^{(t)}, M_{BS(n)})$~\COMMENT{Update the $n$th chunk in $M_{S}$ using the $n$th chunk in $M_{BS}$}
\ENDFOR
\RETURN $M_{S}^{(t+1)}$~\COMMENT{$M_S^{(t)}$ is updated into $M_S^{(t+1)}$}
\end{algorithmic}
\end{algorithm*}

\section{CGO with Ants for Solving the TSP}
\label{sec:cgo-as}

We will realize both MMAS and CGO-AS in the framework of CGO. The fulfillment can provide us an easy way to identify the similarity and the difference between the two ant systems. 

For the TSP, a simple static search landscape $F_R=\langle S, R_{M}, AUX\rangle$ can be easily defined. Each possible tour $\bm{ \pi}$ is a natural state in the problem space $S$. The $R_{M}$ rule can be realized using the tour length $f(\bm{ \pi})$: $\forall \bm{ \pi}_{(a)} ,\bm{ \pi}_{(b)} \in S$, $R_{M}( \bm{ \pi}_{(a)} ,\bm{ \pi}_{(b)})$ returns TRUE if and only if $f(\bm{ \pi}_{(a)})<f(\bm{ \pi}_{(b)})$. $AUX$ simply contains the cost matrix $D=(d_{ij})$.

\subsection{Chunk and Rule Types}
\label{sec:cgo_components}

To implement the algorithms in the framework of CGO, we first define a toolbox of knowledge elements of primary chunk and rule types. We only need a few primary chunk types, including a tour $\bm{ \pi}$, a tour set $\$\bm{ \pi}$, and a pheromone matrix $\varPsi=(\tau_{ij})$.

The following $R_{IE}$ rules are defined. The {\em randomized} $R_{IE}$ rule ($R_{IE}^{RND}$) outputs a tour set $\$\bm{ \pi}$, where each element is randomly generated in $S$. The {\em pheromone matrix} $R_{IE}$ rule ($R_{IE}^{PM}$) outputs a pheromone matrix with $\tau_{ij}=\tau_{max}^{(0)}$, $\forall i,j$. 

The following $R_{UE}$ rules are defined. The {\em greedy} $R_{UE}$ rule, i.e., 
$R_{UE}^{G}(\bm{ \pi}_{(M)},\bm{ \pi}_{(U)})$, $\bm{ \pi}_{(M)}$ is replaced by $\bm{ \pi}_{(U)}$ if and only if 
$R_M(\bm{ \pi}_{(U)},\bm{ \pi}_{(M)})\equiv$TRUE. The {\em pheromone matrix} $R_{UE}$ rule, i.e., $R_{UE}^{PM}$($\varPsi_{(M)}, \$\bm{ \pi}_{(U)}$), updates each $\tau_{ij}\in \varPsi_{(M)}$ by Eq. \ref{Eq:UpdatePheromone}, using the tours in $\$\bm{ \pi}_{(U)}$.

The following $R_{GE}$ rules are defined. The {\em social-only} $R_{GE}$ rule ($R_{GE}^{S}$) takes a pheromone matrix $\varPsi_{(O)}$ as the input, and constructs one tour $\bm{ \pi}_{(C)}$ according to Algorithm \ref{alg:antconstruct}. 

The {\em mixed} $R_{GE}$ rule ($R_{GE}^{M}$) takes the chunk list $\{\bm{ \pi}_{(P)}, \varPsi_{(O)}\}$ as the input, and outputs one tour $\bm{ \pi}_{(C)}$ using Algorithm \ref{alg:antconstruct_mixed}. Algorithm \ref{alg:antconstruct_mixed} has three parameters, $p_{ind}\in [0, 1]$, $\sigma_c \ge 0$, and $w\in [0, 1]$. The proportion $p_c$ is sampled around $p_{ind}$ using a truncated normal distribution (TND) (Line 3), where the underlying normal distribution has $\mu=0$ and $\sigma=\sigma_c$  and lies within the interval $[-w, w]$. In Line 2, $w$ is examined to ensure that $p_c\in[0, 1]$. In Line 4, we define a segment of $\bm{ \pi}_{(P)}$ with $N_P$ edges starting from the $l$th node, where $N_P$ is the nearest integer of $(p_c\cdot N)$, and $l$ is the index of city $i$ in $\bm{ \pi}_{(P)}$. The segment is selected into $\bm{ \pi}_{(C)}$ directly (Lines 6--7). Each remaining city is obtained using $p_{ij}^{(t)}$ (Line 9), the same as Line 3 in Algorithm  \ref{alg:antconstruct}.
Notice that the expectation of $p_c$ is $\mathbb{E}(p_c)=p_{ind}$. Thus the parameter $p_{ind}$ controls the proportion of input information used in $\bm{ \pi}_{(P)}$ and $\varPsi_{(O)}$. If $p_{ind}=0$, Algorithm \ref{alg:antconstruct_mixed} does not use $\bm{ \pi}_{(P)}$, and is equivalent to Algorithm \ref{alg:antconstruct}; If $p_{ind}=1$, there is $\bm{ \pi}_{(C)}=\bm{ \pi}_{(P)}$. Both $\sigma_c$ and $w$ are fixed as 0.1 in this paper. 

\begin{algorithm*} [htb]                     % enter the algorithm environment
\caption{Construct a tour by an ant (agent) using the mixed information $\{\bm{ \pi}_{(P)}, \varPsi_{(O)}\}$}   % give the algorithm a caption
\label{alg:antconstruct_mixed}                           % and a label for \ref{} commands later
\begin{algorithmic}[1]                    % enter the algorithmic environment
\REQUIRE $\{\bm{ \pi}_{(P)}, \varPsi_{(O)}\}$ ~~\COMMENT{The cost matrix $D$ is a default input, and the parameters include $p_{ind}\in [0, 1]$, $\sigma_c \ge 0$, and $w\in [0, 1]$}
\STATE $\mathcal{N}=\{1, \cdots, N\}$; $i=\text{RND($\mathcal{N}$)}$; $\mathcal{N}_i=\mathcal{N}\backslash\{i\}$;  $\pi_{[1]}=i$~~\COMMENT{Start from a randomly selected city}
\STATE {\bf if}~$p_{ind}-w\le 0$~{\bf then}~$w=p_{ind}$;  {\bf if}~$p_{ind}+w\ge 1$~{\bf then}~$w=1-p_{ind}$  ~~\COMMENT{Preprocess $w$ to ensure a valid value} 
\STATE $p_c=p_{ind}+\text{SampleTND}(0, \sigma_c, -w, w)$ ~~\COMMENT{Sample from the truncated normal distribution (TND) within $(-w, w)$}
\STATE $N_P=\text{Round}(p_c\cdot N)$; $l=\text{GetIndex}(i, \bm{ \pi}_{(P)})$~\COMMENT{Define a segment of $\bm{ \pi}_{(P)}$ with $N_P$ edges, starting from the $l$th node} 
\FOR{$n =$ 2 to $N$} 
\IF {$n \le N_P+1$}
\STATE $j=\pi_{(P)[l]}$; $l=l+1$; {\bf if}~$l>N$~{\bf then}~$l=1$ \COMMENT{Select the $l$th node of $\bm{ \pi}_{(P)}$ and go to the next node of $\bm{ \pi}_{(P)}$}
\ELSE
\STATE Select the next city $j\in \mathcal{N}_i$ with the probability $p_{ij}^{(t)}$, using Eq. \ref{Eq:SelProb} with $\varPsi_{(O)}$ and $D$
\ENDIF
\STATE $\pi_{[n]}=j$; $\mathcal{N}_j=\mathcal{N}_i\backslash\{j\}$; $i=j$~~~~~~~~~~~~~~~~~~~~~~~~~~~~~~~~~~~~~~~~~~~~~\COMMENT{Move to the next nodes $j$} 
\ENDFOR
\RETURN $\bm{ \pi}_{(C)}=(\pi_{[1]}, \cdots, \pi_{[N]})$ ~~~~~~~~~~~~~~~~~~~~~~~~~~~~~~~~~~~~~~~~~~~~~~\COMMENT{Return the new tour}
\end{algorithmic}
\end{algorithm*}

The {\em 3-opt} $R_{GE}$ rule ($R_{GE}^{3O}$) is the same as the 3-opt local search algorithm used in~\cite{Stutzle:2000p2978}. The 3-opt local search algorithm proceeds by systematically testing the incumbent tour $\bm{ \pi}_{(C)}$, with some standard speed-up techniques using nearest neighbors \cite{bentley1992fast,Stutzle:2000p2978}, and with the technique of {\em don't look bits} on each node \cite{bentley1992fast}. Notice that the tour $\bm{ \pi}_{(C)}$ can be improved by replacing at most three edges in each test. By default, the candidate list at length 20 is considered \cite{Stutzle:2000p2978}.  

The macro $R_{GE}$ rule $R_{GE}^{S+3O}$  is defined as the tuple $\langle R_{GE}^{S}, R_{GE}^{3O} \rangle$, where the input is $\varPsi_{(O)}$, and the output of $R_{GE}^{S}$ is further processed by $R_{GE}^{3O}$. 

The macro $R_{GE}$ rule $R_{GE}^{M+3O}$ is defined as the tuple $\langle R_{GE}^{M}, R_{GE}^{3O} \rangle$, where the input is $\{\bm{ \pi}_{(P)}, \varPsi_{(O)}\}$, and the output of $R_{GE}^{M}$ further processed by $R_{GE}^{3O}$.

\subsection{SPEC-MP and ESHs}

For MMAS, the memory elements are defined as follows: $M_A$ is empty, $M_S$ contains the pheromone matrix $\varPsi$, and $M_G$ contains one tour $\bm{ \pi}_{C}$. For CGO-AS, the memory elements are defined as follows: $M_A$ contains one tour $\bm{ \pi}_{P}$, $M_S$ contains the pheromone matrix $\varPsi$, and $M_G$ contains one tour $\bm{ \pi}_{C}$. 

Both MMAS and CGO-AS can work on the same SPEC-MP, as defined in Table \ref{tab:SPEC-MP}. The difference between them is that MMAS only uses $\varPsi$ in $M_S$, whereas CGO-AS also uses $\bm{ \pi}_{P}$ in $M_A$. Normally, some additional knowledge might be embedded with the SPEC-MP. For example, as $R_{UE}^{G}$ is used, $\bm{ \pi}_{P}$ in $M_A$ always retains the personal best solution for an agent.

\begin{table*} [!ht]
\small
\renewcommand{\arraystretch}{1.2}
\centering \caption{The memory protocol rows in SPEC-MP for CGO-AS}
\begin{tabular}{|c|c|c|c|c|} \hline 
$ID_{M}$ & $CH_{M}$ & $R_{IE}$ Instance & $R_{UE}$ Instance & $CH_{U}$ \\ \hline 
 $M_{A}$ & $\bm{ \pi}_{P}$ & \small $R_{IE}^{RND}$ & \small $R_{UE}^{G}$ & $\bm{ \pi}_{C}$\\ \hline  
 $M_{S}$ & $\varPsi$ & \small $R_{IE}^{PM}$ & \small $R_{UE}^{PM}$ & $\bm{ \pi}_{C} $ \\ \hline 
\end{tabular}
\label{tab:SPEC-MP}
\end{table*}

Table \ref{tab:SPEC-G} lists the embedded search heuristics (ESHs) of this work, including MMAS, MMAS$_{3opt}$, CGO-AS, and CGO-AS$_{3opt}$, where MMAS$_{3opt}$ and CGO-AS$_{3opt}$ also apply the 3-opt local search. Note that the $R_{GE}^{M}$ rules in CGO-AS and CGO-AS$_{3opt}$ both have a setting parameter $p_{ind} \in [0, 1]$ in Algorithm \ref{alg:antconstruct_mixed}, and are respectively reduced to MMAS and  MMAS$_{3opt}$ if $p_{ind}=0$.

\begin{table*} [!ht]
\renewcommand{\arraystretch}{1.2}
\small
\centering \caption{The list of embedded search heuristics (ESHs)}
\begin{tabular}{|c|c|c|c|} \hline 
ESH & $R_{GE}$ Instance & $E_{IG}$ & $CH_{OG}$ \\ \hline 
MMAS & \small $R_{GE}^{S}$ & \{$\varPsi$\} & $\bm{ \pi}_{C}$ \\ \hline 
MMAS$_{3opt}$ & \small $R_{GE}^{S+3O}$ & \{$\varPsi$\} & $\bm{ \pi}_{C}$ \\ \hline 
CGO-AS & \small $R_{GE}^{M}$ & \{$\bm{ \pi}_{P},\varPsi$\} & $\bm{ \pi}_{C}$  \\ \hline 
CGO-AS$_{3opt}$ & \small $R_{GE}^{M+3O}$ & \{$\bm{ \pi}_{P},\varPsi$\} & $\bm{ \pi}_{C}$  \\ \hline 
\end{tabular}
\label{tab:SPEC-G}
\end{table*}

\subsection{Brief Summary}

Although CGO looks a little bit complex, the realization of algorithms in this framework is smooth. CGO-AS can be reduced into an implementation of algorithmic components (chunks and rules in Section \ref{sec:cgo_components}) with a simple script description (SPEC-MP and ESHs in Tables \ref{tab:SPEC-MP} and \ref{tab:SPEC-G}). 

We can easily identify the similarities and differences between MMAS and CGO-AS, base on the framework of CGO as shown in Figure \ref{Fig:cgos-as-01}. Each ant is represented as an agent in CGO. The pheromone matrix is stored in the social memory $M_S$, and it is updated using $M_{BS}$ and $B_{US}$. There are two main differences. First, CGO-AS uses additional modules including the individual memory $M_A$ and its maintenance modules $M_{BA}$ and $B_{UA}$. Second, in $B_{GEN}$, the elemental generating rule is changed from Algorithm \ref{alg:antconstruct} (in MMAS), which only uses pheromone trails, to Algorithm \ref{alg:antconstruct_mixed} (in CGO-AS), which uses both individual and social learning with the proportion controlled by $p_{ind}$. 

With the two differences, the memory form is transformed from one single social memory into one social plus multiple individual memories, and the learning form is upgraded from a pure social learning into a mixed social and individual learning.  The framework of CGO provides a support for the self-organized interactions between multiple memories and behaviors.

\section{Results and Discussion}
\label{sec:results}

We conduct a series of experiments to evaluate the performance of two algorithms, CGO-AS$_{3opt}$ and MMAS$_{3opt}$ that incorporates the 3-opt local search rule $R_{GE}^{3O}$. Due to an application of the 3-opt and other local search heuristic \footnote{Note that the LK heuristic \cite{lin1973effective} and its variants \cite{Applegate:2003p2830,Helsgaun:2000p2779} are often considered as the best performing local search strategies, with respect to the solution quality. Nevertheless, the LK heuristic is more different to implement than 2-opt and 3-opt, and it requires careful fine-tuning to run efficiently \cite{Stutzle:2000p2978}.}, local optima of the search landscape of TSP exhibits a ``big valley'' structure \cite{Stutzle:2000p2978,Merz:2001p2780}. This suggests that the high-quality tours tend to concentrate on a very small subspace around the optimal tour(s). On the basic group parameters, we consider $T=500$, and $K \in \{10, 30, 50\}$. For the 3-opt local search, 20 nearest neighbors are used. On the other parameters, we consider default settings, including $\alpha=1$, $\beta=2$, and $p_{\text{best}}=0.05$. In MMAS \cite{Stutzle:2000p2978}, a schedule is used to alternate the pheromone-trail update between $\bm{ \pi}^{(t)}_{\text{ib}}$ and $\bm{ \pi}^{(t)}_{\text{gb}}$. In the first 25, 75, 125, 250, and the later iterations, the intervals using $\bm{ \pi}^{(t)}_{\text{gb}}$ are respectively 25, 5, 3, 2, and 1.   

\begin{figure*} [tbh]
\centering \includegraphics[width=4.20in]{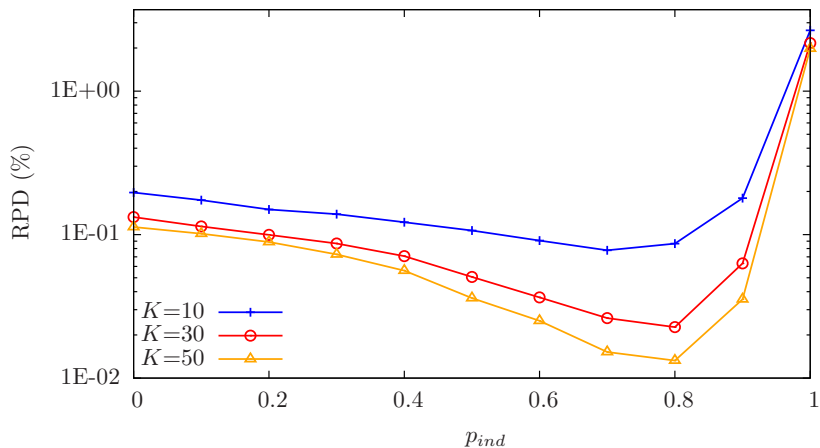} \caption{Average RPD of mean results by CGO-AS$_{3opt}$ with $K$=\{10, 30, 50\} and $p_{ind}\in [0, 1]$.}
\label{Fig:CGO-AS_3OPT_QS}
\end{figure*}

We perform experiments on the widely-used benchmark instances in TSPLIB \cite{Reinelt:1991p2617}. For each problem instance, 100 independent runs have been performed for obtaining the mean results in statistic. For each mean result $f$, we report $100 \cdot (f-f^*)/f^*$, the relative percentage deviation (RPD) as an evaluation value of $f$, and $100 \cdot \sigma/f^*$, as an evaluation value of the standard deviation (SD) $\sigma$, where $f^*$ is the the optimal value. The minimal RPD of the optimal solution is 0.

\begin{figure} [thb]
\centering
\begin{subfigure}{.5\textwidth}
  \centering
  \includegraphics[width=1\linewidth]{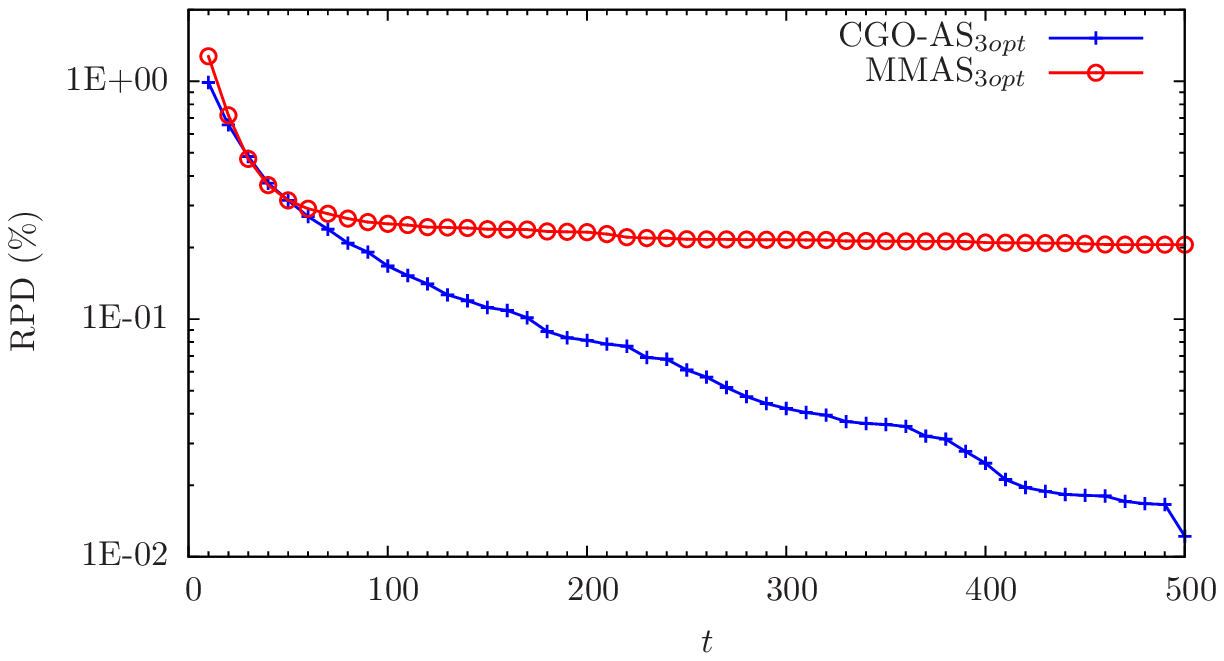} 
  \caption{pcb442}
  \label{Fig:converge_pcb442}
\end{subfigure}%
\begin{subfigure}{.5\textwidth}
  \centering
  \includegraphics[width=1\linewidth]{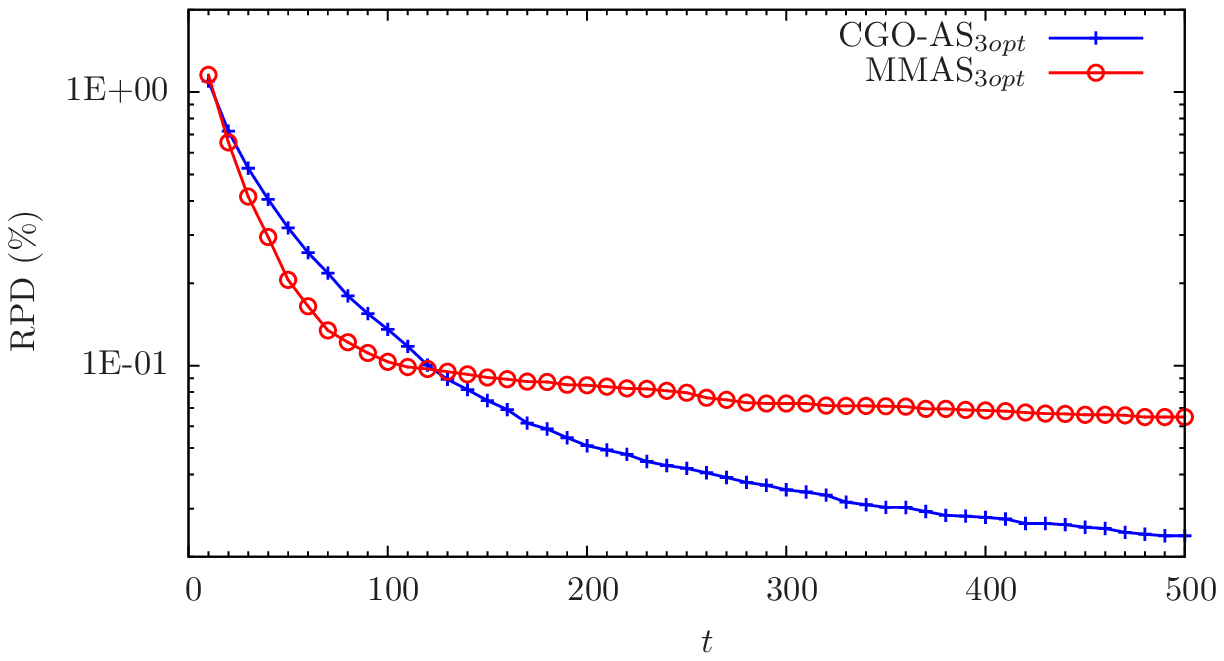} 
  \caption{att532}
  \label{Fig:converge_att532}
\end{subfigure}

\begin{subfigure}{.5\textwidth}
  \centering
  \includegraphics[width=1\linewidth]{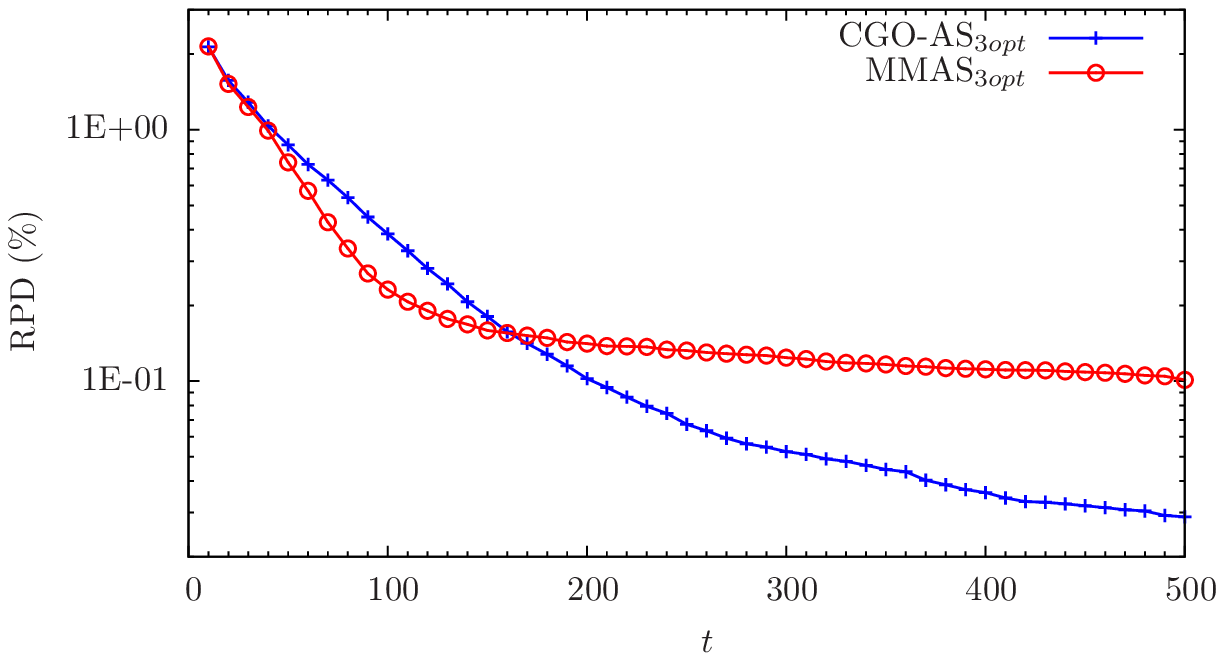} 
  \caption{rat783}
  \label{Fig:converge_rat783}
\end{subfigure}%
\begin{subfigure}{.5\textwidth}
  \centering
  \includegraphics[width=1\linewidth]{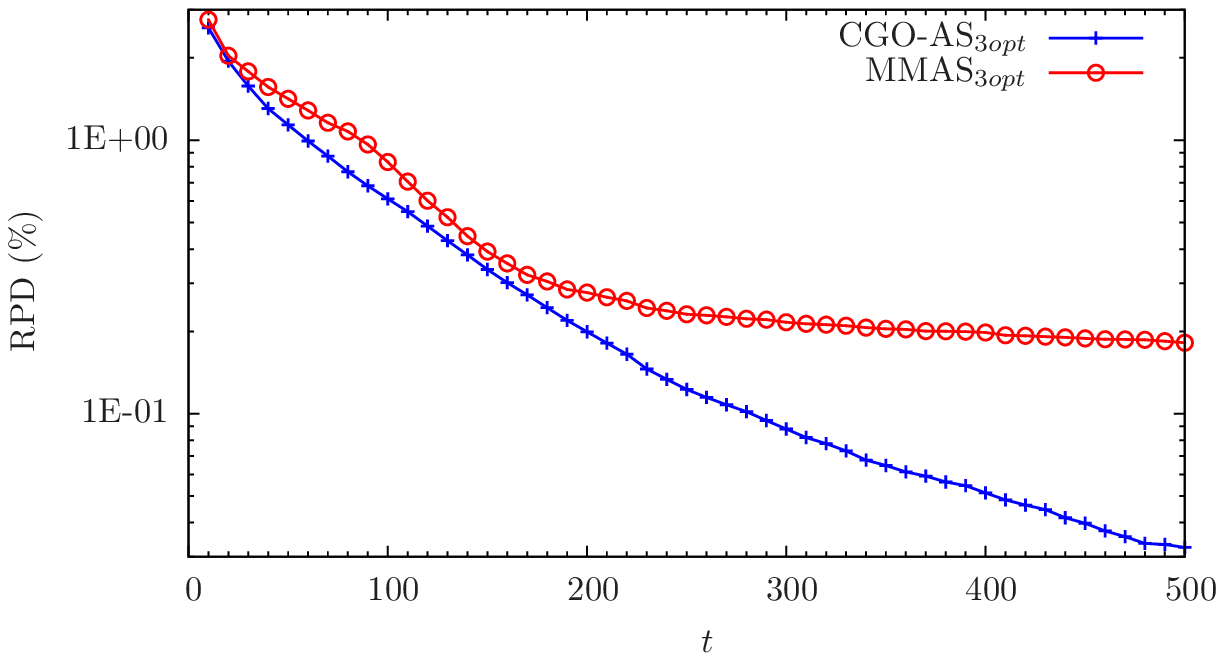} 
  \caption{pcb1173}
  \label{Fig:converge_pcb1173}
\end{subfigure}

\begin{subfigure}{.5\textwidth}
  \centering
  \includegraphics[width=1\linewidth]{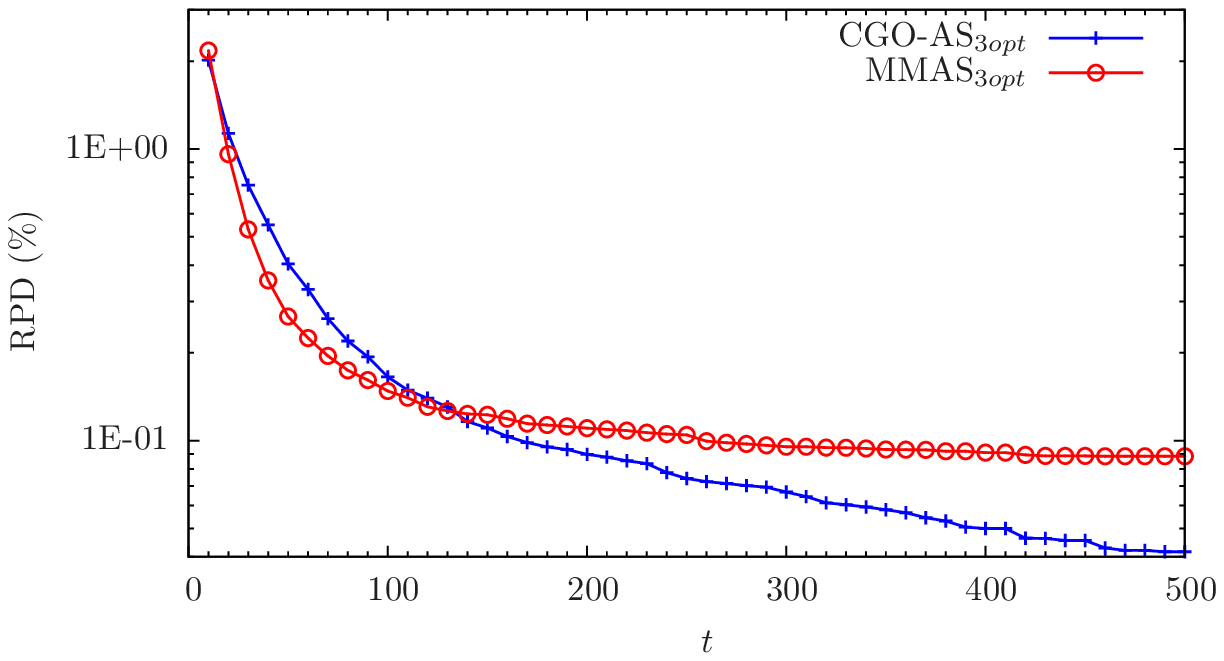} 
  \caption{d1291}
  \label{Fig:converge_d1291}
\end{subfigure}%
\begin{subfigure}{.5\textwidth}
  \centering
  \includegraphics[width=1\linewidth]{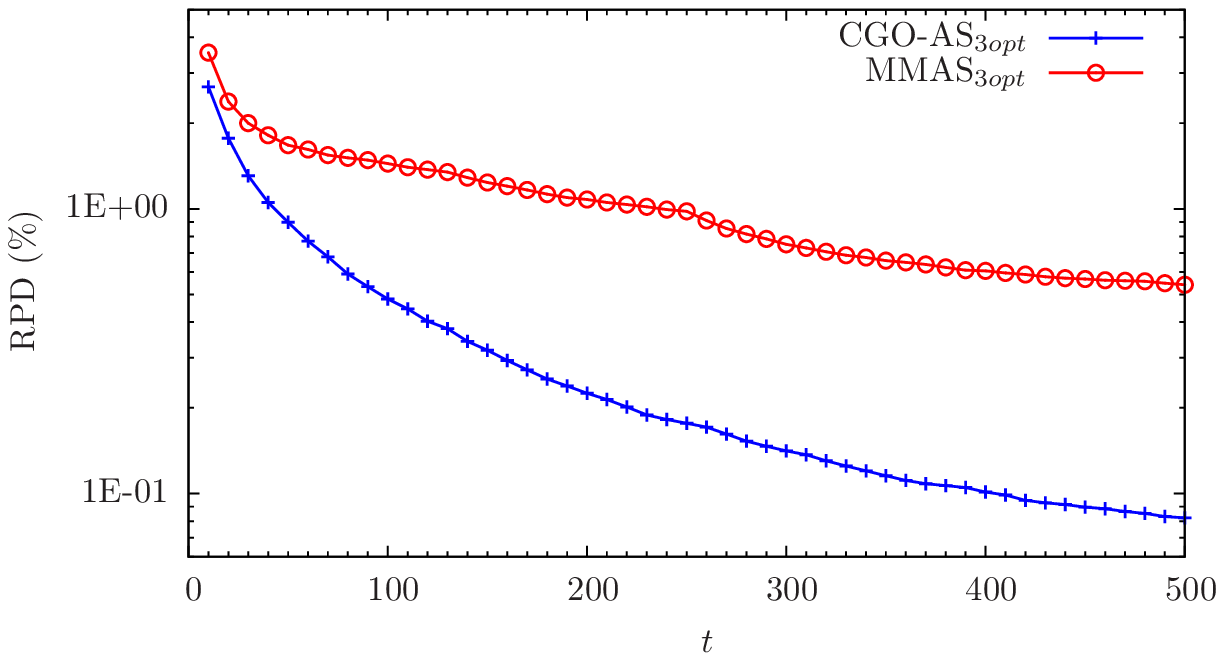} 
  \caption{fl1577}
  \label{Fig:converge_fl1577}
\end{subfigure}
\caption{RPD of mean results by CGO-AS$_{3opt}$ and MMAS$_{3opt}$ with $K$=30 over the 500 learning cycles.}
\label{fig:cgo_convergence}
\end{figure}

We first evaluate the performance on 10 benchmark instances in the range of $N$ from 51 to 1577 (as listed in Tables \ref{tab:CompMMAS} (a)--(c) of Appendix \ref{sec:app1}), which are a set of instances frequently used in the literature. Figure \ref{Fig:CGO-AS_3OPT_QS} shows the average RPD of the mean by CGO-AS$_{3opt}$ with $K$=\{10, 30, 50\}, for $p_{ind}$ in the range from 0 to 1. For the agents in CGO-AS, $p_{ind}$ controls the proportion using knowledge between individual and social memory, i.e. the weighting balance between individual and social learning. As shown in the figure, CGO-AS$_{3opt}$ cannot reach a good performance at either end, i.e., in the case of $p_{ind} = 0$ or $p_{ind} = 1$. In the case of $p_{ind} = 0$, ants only use social memory (pheromone trails), for which MMAS$_{3opt}$ is in fact this special case of CGO-AS$_{3opt}$. In the case of $p_{ind} = 1$, ants only use their own individual (route) memories and perform local searching. Interestingly, the results showed that CGO-AS$_{3opt}$ reached a much better performance when ants take a mixed usage combining both individual and social memory together. In this case, ants uses two parts of learning. One part of learning uses social memory that contains an accumulated adaptive knowledge~\cite{Boyd2011} for accelerating the learning process. Another part of learning uses individual route memory that preserves novel patterns~\cite{Nemeth:1986p980} for supporting the capability to escape from some maladaptive outcomes~\cite{Boyd2011}. 
As shown in Figure \ref{Fig:CGO-AS_3OPT_QS}, CGO-AS$_{3opt}$ reaches the best performance of searching at around $p_{ind} = 0.8$. This means that the best searching relies more on individual memory, which is called as socially biased individual learning (SBIL) in the field of animal learning \cite{Galef:1995p1128}. The result is interpretative from the viewpoint of searching for optimization. Each ant in CGO-AS performs a local searching based on its individual memory; at the same time, it also efficiently search ``big valleys'' of TSP landscape with the guide of the high-quality heuristic cues in the social memory accumulated by all ants (simulating the pheromone trails of natural ants). If ants only use individual information, i.e. in the case of $p_{ind} = 1$, the search would more likely be trapped to the local minima. In contrast, if ants only use social information (such as the pheromone matrix used by the existing ant systems), i.e. in the case of $p_{ind} = 0$, it is challenging to adaptively maintain a diversity in searching. %, leading to a difficulty in preventing the search from a premature convergence. 
Aiming to prevent the search of AS from a premature convergence, previous research on MMAS \cite{Stutzle:2000p2978} had attempted to introduce the mechanism linking diversity into pheromone trail by re-initialization, but showed a very limited success. 

More detailed results on the 10 instances are provided in Appendix \ref{sec:app1}. For each tested instance, Tables \ref{tab:CompMMAS} (a)--(c) give the ratios of the runs reaching the optimal solution (Best), the RPD of the mean values (Mean), and the standard deviations (SD) by CGO-AS$_{3opt}$ (with $p_{ind}=0.8$) and MMAS$_{3opt}$ for the experiments with $K \in \{10, 30, 50\}$ respectively. The results show that CGO-AS$_{3opt}$ has achieved a significant better performance than MMAS$_{3opt}$. In comparison with MMAS$_{3opt}$, for all the instances, CGO-AS$_{3opt}$ gains a much bigger ratio of the runs reaching the optimal solution, a smaller RPD of the mean value, and a lower standard deviation, see Tables \ref{tab:CompMMAS} (a)--(c).  CGO-AS$_{3opt}$ with $K=10$ outperforms MMAS$_{3opt}$ with $K=50$ on the RPD of the mean value and the standard deviation, and beats MMAS$_{3opt}$ with $K=30$ completely on all the resulting values. In the case with $K=50$, CGO-AS$_{3opt}$ is able to solve two more instances (d198 and lin318) than MMAS$_{3opt}$ in all runs, and to find the optimal solutions for all instances including fl1577.

To observe more information for understanding the learning process more clearly, we show Figure \ref{fig:cgo_convergence}, the RPD results of CGO-AS$_{3opt}$ (with $p_{ind} = 0.8$) and MMAS$_{3opt}$ for six larger TSP instances with $K$=30 over 500 learning cycles. For MMAS$_{3opt}$ (which only uses social memory), its learning process quickly stagnated at the local and lower-quality minima, although it holds a fast learning speed in its early learning cycles. For all the tested instances here, CGO-AS$_{3opt}$ outperforms MMAS$_{3opt}$ to reach a much better quality in solution. It is interesting that although MMAS$_{3opt}$ has a quicker learning speed at the very beginning learning cycles, CGO-AS quickly catches up the learning speed of MMAS$_{3opt}$, and keeps a much faster learning pace than MMAS$_{3opt}$ in the following later learning cycles, and finally reaches a better solution. For some tests (such as f1577), CGO-AS has a faster learning speed than MMAS$_{3opt}$ in the whole learning cycles.%, even at the very beginning.

\begin{figure} [thb]
\centering
\begin{subfigure}{.5\textwidth}
  \centering
  \includegraphics[width=.95\linewidth]{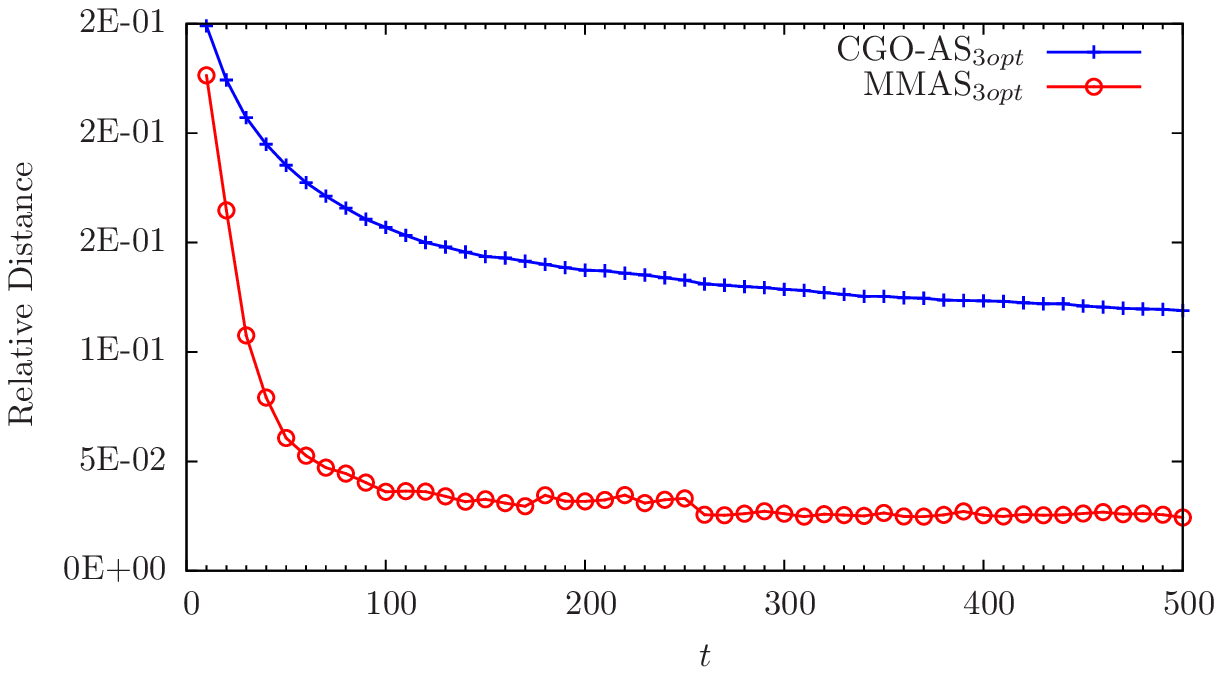} 
  \caption{pcb442}
  \label{Fig:distance_pcb442}
\end{subfigure}%
\begin{subfigure}{.5\textwidth}
  \centering
  \includegraphics[width=.95\linewidth]{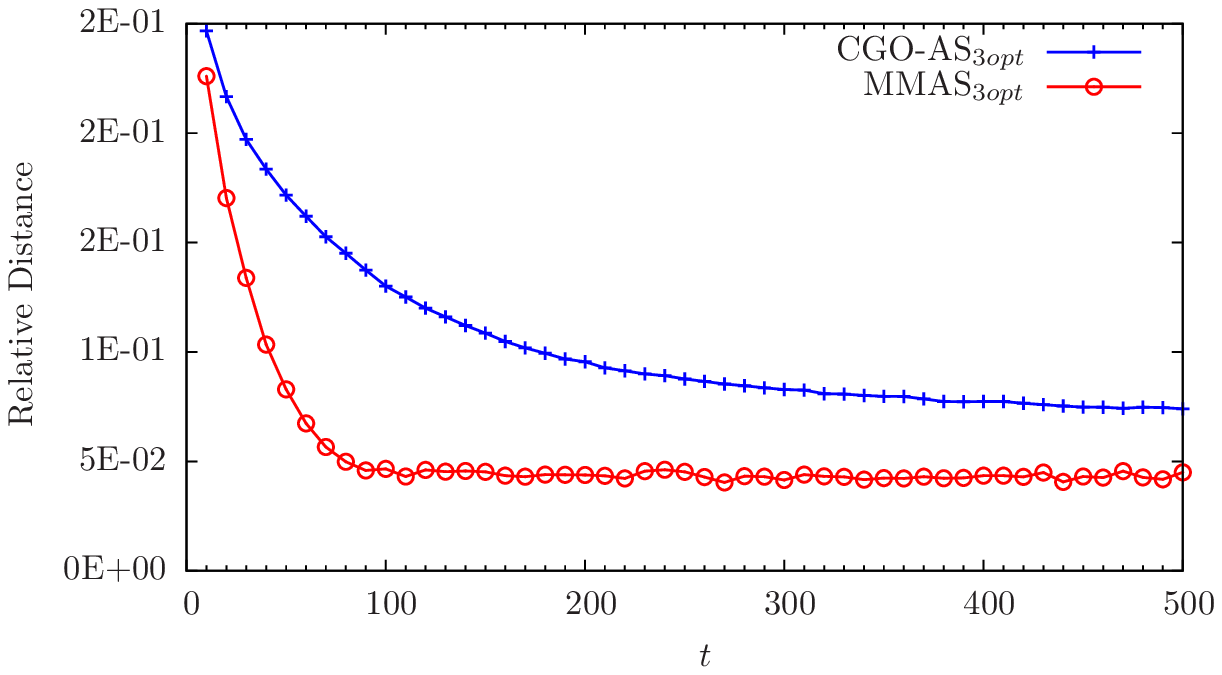} 
  \caption{att532}
  \label{Fig:distance_att532}
\end{subfigure}

\begin{subfigure}{.5\textwidth}
  \centering
  \includegraphics[width=.95\linewidth]{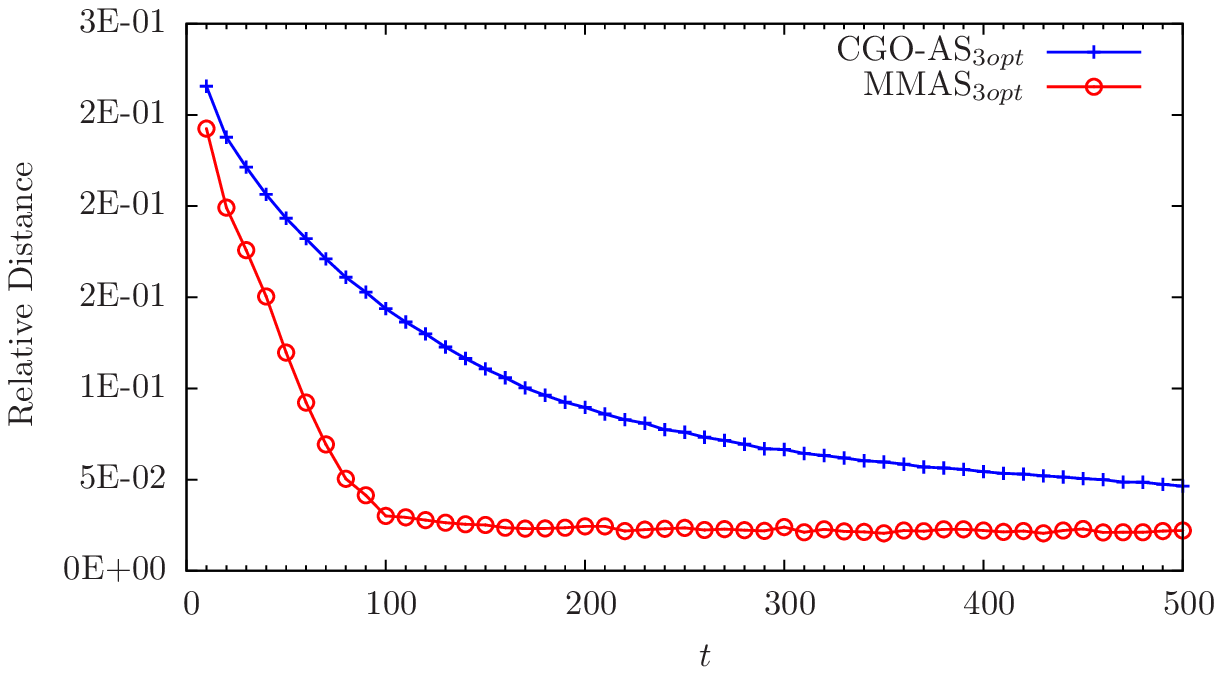} 
  \caption{rat783}
  \label{Fig:distance_rat783}
\end{subfigure}%
\begin{subfigure}{.5\textwidth}
  \centering
  \includegraphics[width=.95\linewidth]{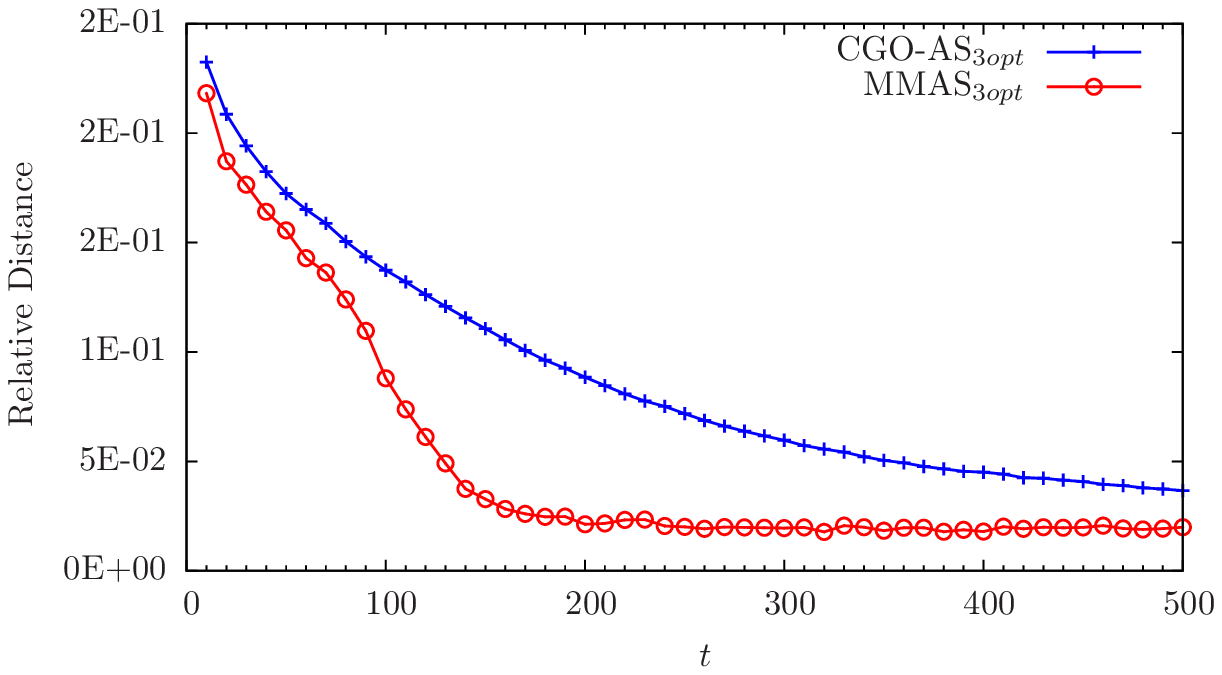} 
  \caption{pcb1173}
  \label{Fig:distance_pcb1173}
\end{subfigure}

\begin{subfigure}{.5\textwidth}
  \centering
  \includegraphics[width=.95\linewidth]{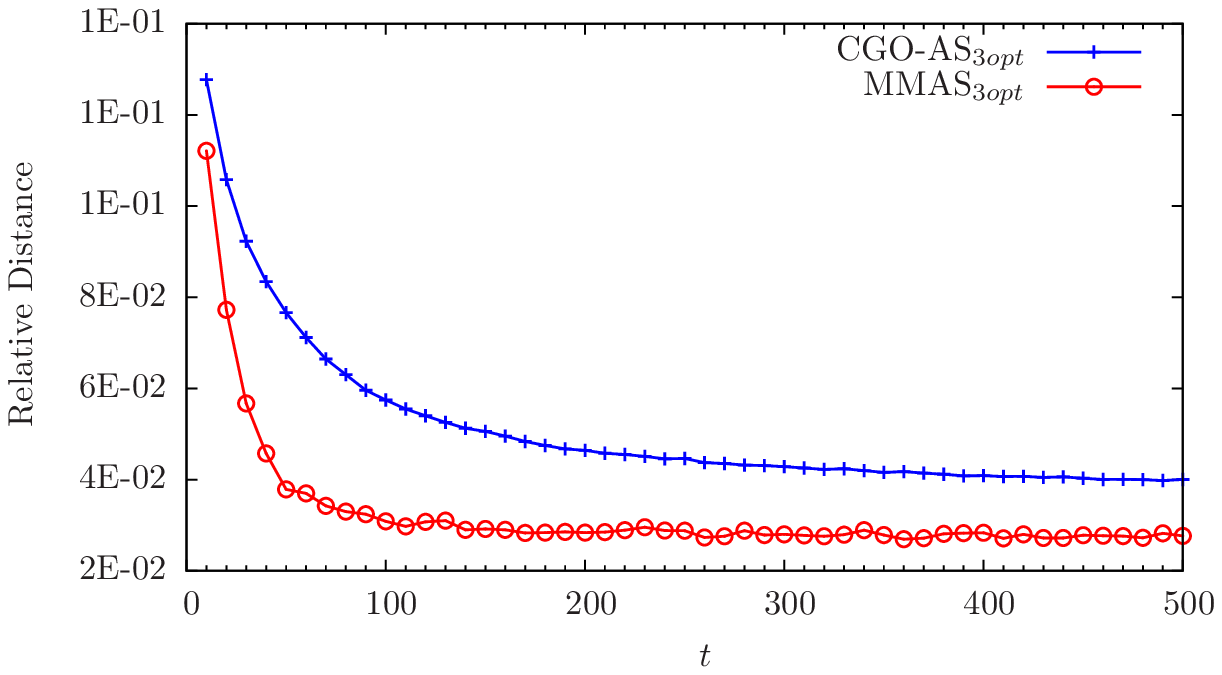} 
  \caption{d1291}
  \label{Fig:distance_d1291}
\end{subfigure}%
\begin{subfigure}{.5\textwidth}
  \centering
  \includegraphics[width=.95\linewidth]{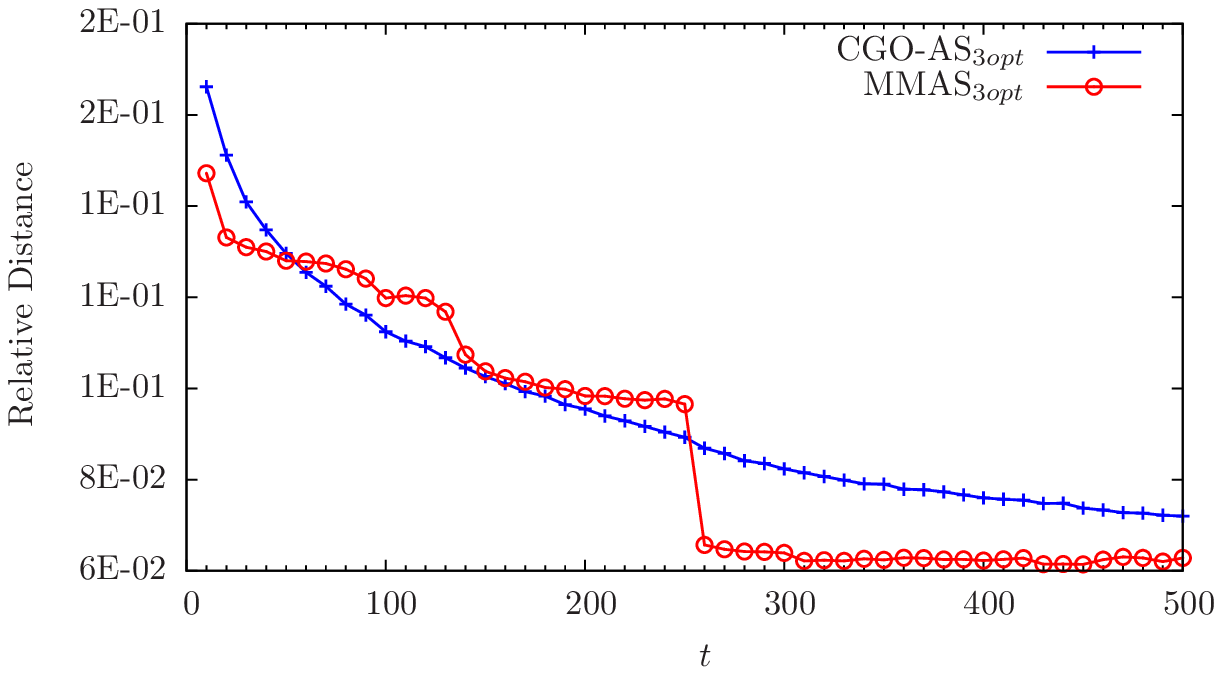} 
  \caption{fl1577}
  \label{Fig:distance_fl1577}
\end{subfigure}
\caption{Population diversity by CGO-AS$_{3opt}$ and MMAS$_{3opt}$ with $K$=30 over the 500 learning cycles.}
\label{fig:cgo_distancence}
\end{figure}

Figure \ref{fig:cgo_distancence} gives the comparison in population diversity between CGO-AS$_{3opt}$ (with $p_{ind} = 0.8$) and MMAS$_{3opt}$, by showing the relative distances between $\{\bm{ \pi}_{C(k)}^{(t)}| k\in [1,K]\}$ (i.e., the set of newly generated states by the agents) of CGO-AS$_{3opt}$ and MMAS$_{3opt}$ for six larger TSP instances with $K$=30 over 500 learning cycles. The relative distance is defined as $\bar{d}/N$, where $\bar{d}=\sum_{k_1<k_2} d(\bm{ \pi}_{C(k_1)}^{(t)}, \bm{ \pi}_{C(k_2)}^{(t)})/C_2^K$, and the distance $d$ between any two states is given by the number of different edges. As shown in Figure~\ref{fig:cgo_distancence}, CGO-AS$_{3opt}$ holds a higher diversity in population than MMAS$_{3opt}$  over the learning cycles for almost all the test instances. The population diversity among the agents in CGO-AS$_{3opt}$ is adaptively maintained by their individual memory in the learning process. Each agent maintains the personal best state $\bm{ \pi}_{P}$ in its $M_A$ (based on the specification in Table \ref{tab:SPEC-MP}), and generates each new state largely inheriting from the high-quality information in its $M_A$, as $p_{ind}$ in Algorithm \ref{alg:antconstruct_mixed} is sufficiently large. Holding population diversity is important for optimization, as it has been shown to play a significant role of effectiveness in the problem solving of human groups \cite{Paulus:2000p1114,kavadias2009effects}. 

\begin{table*} [htb]
\small
\centering \caption{Results by CGO-AS$_{3opt}$ with $K$=10, PSO-ACO-3Opt, and FOGS-ACO.}
\begin{tabular}{|l|c|c|c|c|c|c|c|c|} \hline 
    \multirow{2}{*}{Instance} &
    \multirow{2}{*}{$f_{opt}$} &
      \multicolumn{3}{|c|}{CGO-AS$_{3opt}$} &
      \multicolumn{2}{|c|}{PSO-ACO-3Opt} &
      \multicolumn{2}{|c|}{FOGS-ACO} \\ \cline{3-9}

 & & Mean (\%) & SD (\%) & Time & Mean (\%) & SD (\%) & Mean (\%) & SD (\%) \\ \hline 
eil51    & 426 & 0.000 & 0.0000 & 0.002 & 0.106 & 0.1432 & 2.406 & 1.2465  \\ \hline 
berlin52 & 7542 & 0.000 & 0.0000 & 0.001 & 0.016 & 0.0314 & 0.526 & 0.6310  \\ \hline 
st70     & 675 & 0.000 & 0.0000 & 0.003 & 0.474 & 0.2178 & 2.873 & 1.2356  \\ \hline 
eil76    & 538 & 0.000 & 0.0000 & 0.005 & 0.056 & 0.0874 & 1.976 & 1.0762  \\ \hline 
pr76     & 108159 & 0.000 & 0.0000 & 0.002 & - & - & 2.522 & 1.8274  \\ \hline 
rat99    & 1211 & 0.000 & 0.0000 & 0.013 & 1.354 & 0.1635 & - & -  \\ \hline 
kroA100  & 21282 & 0.000 & 0.0000 & 0.003 & 0.766 & 0.3676 & 0.682 & 2.9804  \\ \hline 
rd100    & 7910 & 0.000 & 0.0000 & 0.005 & - & - & 2.248 & 1.1876  \\ \hline 
eil101   & 629 & 0.000 & 0.0000 & 0.014 & 0.588 & 0.3370 & 3.919 & 3.4118  \\ \hline 
lin105   & 14379 & 0.000 & 0.0000 & 0.002 & 0.001 & 0.0033 & - & -  \\ \hline 
ch150    & 6528 & 0.000 & 0.0000 & 0.029 & 0.551 & 0.4225 & - & -  \\ \hline 
kroA200  & 29368 & 0.000 & 0.0000 & 0.057 & 0.947 & 0.3906 & 7.314 & 2.2953  \\ \hline 
\end{tabular}
\label{tab:TestTSP200}
\end{table*}

\begin{table*} [htb]
\small
\centering \caption{Results by CGO-AS$_{3opt}$ with $K$=50, DIWO, and DCS.}
\begin{tabular}{|l|c|c|c|c|c|c|c|c|} \hline 
    \multirow{2}{*}{Instance} &
    \multirow{2}{*}{$f_{opt}$} &
      \multicolumn{3}{|c|}{CGO-AS$_{3opt}$} &
      \multicolumn{2}{|c|}{DIWO} &
      \multicolumn{2}{|c|}{DCS} \\ \cline{3-9}

 & & Mean (\%) & SD (\%) & Time & Mean (\%) & SD (\%) & Mean (\%) & SD (\%) \\ \hline 
tsp225 & 3916 & 0.000 & 0.000 & 0.16 & 2.395 & 2.821 & 1.092 & 0.529  \\ \hline 
pr226  & 80369 & 0.000 & 0.000 & 0.07 & 0.224 & 0.192 & 0.022 & 0.075  \\ \hline 
pr264  & 49135 & 0.000 & 0.000 & 0.05 & - & - & 0.249 & 0.326  \\ \hline 
a280   & 2579 & 0.000 & 0.000 & 0.07 & 0.768 & 0.700 & 0.517 & 0.460  \\ \hline 
pr299  & 48191 & 0.000 & 0.000 & 0.20 & - & - & 0.580 & 0.273  \\ \hline 
lin318 & 42029 & 0.000 & 0.000 & 0.84 & - & - & 0.965 & 0.441  \\ \hline 
rd400  & 15281 & 0.000 & 0.001 & 1.81 & 2.423 & 0.863 & 1.654 & 0.396  \\ \hline 
fl417  & 11861 & 0.000 & 0.000 & 0.26 & - & - & 0.418 & 0.172  \\ \hline 
pcb442 & 50778 & 0.005 & 0.027 & 1.81 & 2.173 & 0.799 & - & -  \\ \hline 
pr439  & 107217 & 0.000 & 0.003 & 1.23 & - & - & 0.693 & 0.409  \\ \hline 
att532 & 27686 & 0.013 & 0.021 & 4.60 & 1.874 & 0.802 & - & -  \\ \hline 
rat575 & 6773 & 0.023 & 0.019 & 4.46 & - & - & 2.713 & 0.528  \\ \hline 
rat783 & 8806 & 0.012 & 0.017 & 8.74 & - & - & 3.444 & 0.433  \\ \hline 
pr1002 & 259045 & 0.028 & 0.034 & 12.81 & 3.187 & 2.604 & 3.700 & 0.435  \\ \hline 
\end{tabular}
\label{tab:TestTSP500}
\end{table*}

Next, we evaluate the performance of CGO-AS$_{3opt}$ by comparing it with other ant systems and some recently published algorithms, in terms of the RPD of mean results and standard deviations in optimization respective to the best solutions. We run CGO-AS$_{3opt}$ on AMD Phenom II 3.4 GHz, and report the computational speed of CGO-AS$_{3opt}$ with the CPU time (in seconds). In Tables \ref{tab:TestTSP200} and \ref{tab:TestTSP500}, the symbol ``-'' means that no result was provided in the references.

Table \ref{tab:TestTSP200} gives the comparison among CGO-AS$_{3opt}$ with $K=10$ and two ant systems, PSO-ACO-3Opt \cite{mahi2015new} and FOGS-ACO \cite{saenphon2014combining}. PSO-ACO-3Opt is an ACO algorithm with a set of performance parameters that are optimized using both PSO and the 3-opt local search operator. FOGS-ACO is a hybrid algorithm of the Fast Opposite Gradient Search (FOGS) and ACO. The test is performed on a set of TSPLIB instances, where the number of nodes is from 51 to 200 used by the two ant systems. As shown in Table~\ref{tab:TestTSP200}, for all the instances, CGO-AS$_{3opt}$ reaches the optimal value in all runs, and outperforms both PSO-ACO-3Opt and FOGS-ACO. 

Table \ref{tab:TestTSP500} gives the comparison between CGO-AS$_{3opt}$ with $K=50$ and two other recently published optimization algorithms, DIWO \cite{zhou2015discrete} and DCS \cite{ouaarab2014discrete}. DIWO is a discrete invasive weed optimization (IWO) algorithm with the 3-opt local search operator. DCS is a discrete cuckoo search algorithm for solving the TSP, which can reconstructs its population to introduce a new category of cuckoos. The test is performed on a set of TSPLIB instances, where the number of nodes is from 225 to 1002 used by the two optimization algorithms. As shown in Table \ref{tab:TestTSP500}, for all the instances, CGO-AS$_{3opt}$ again outperforms both DIWO and FOGS-ACO. CGO-AS$_{3opt}$ reaches the optimal value for seven of the test instances in all runs, and approaches to the near optimal value for the other test instances.

\section{Conclusion}
\label{sec:Conclusion}

In this paper, we presented CGO-AS, a generalized ant system that is implemented in the framework of cooperative group optimization. AS is an algorithm simulating the foraging system that uses pheromone trails in ants colonies. CGO is a framework to support the cooperative search process by a group of agents. CGO-AS has combined and used both individual route memory and social pheromone trails to simulate the intelligence of natural ants, therefore it enables us to leverage the power of mixed individual and social learning. We have not attempted to provide a complete comparison in performance between CGO-AS and the existing algorithms on various problems. Rather, we aim at showing the benefit of using mixed social and individual learning in optimization. We tested the performance of CGO-AS for elucidating the weighting balance between individual and social learning, and compared it with the existing AS systems and some recently published algorithms, using the well-known traveling salesman problem (TSP) as a benchmark.  

The results on the instances in TSPLIB showed that the group of agents (ants) with a mixed usage of individual and social memory reaches a much better performance of search than the systems using either individual memory or social memory only. The best performance is gained under the condition when agents use individual memory as their primary information source, and simultaneously also use social memory as their searching guidance. The tests showed that CGO-AS not only reaches a better quality in solution, but also holds a faster learning speed, especially in later learning cycles. The benefit of optimization may be due to the introduced mechanism in the CGO-AS algorithm that adaptively maintains the population diversity using the information learned and stored in the individual memory of each agent, and also accelerates the learning process using the knowledge accumulated in the social memory of agents. The performance of CGO-AS$_{3opt}$ turned out to be competitive in comparison with the existing AS systems and some recent published algorithms, including MMAS$_{3opt}$, PSO-ACO-3Opt, FOGS-ACO, DIWO, and DCS.

\newpage
\begin{appendices}

\section{}
\label{sec:app1}

Tables \ref{tab:CompMMAS} (a)--(c) give the ratios of the runs reaching the optimal solution (Best), the RPD of the mean values (Mean), and the standard deviations (SD) by CGO-AS$_{3opt}$ (with $p_{ind}=0.8$) and MMAS$_{3opt}$ for the experiments with $K \in \{10, 30, 50\}$ respectively, on 10 benchmark instances in the range of $N$ from 51 to 1577, which are a set of instances frequently used in the literature.

\setcounter{table}{0}
\renewcommand{\thetable}{A\arabic{table}}

\begin{table}[H]
\centering
\caption{Results by CGO-AS$_{3opt}$ and MMAS$_{3opt}$ on 10 benchmark instances in the range of $N$ from 51 to 1577.}
\label{tab:CompMMAS}

{\centering \includegraphics[width=0.7507\textwidth]{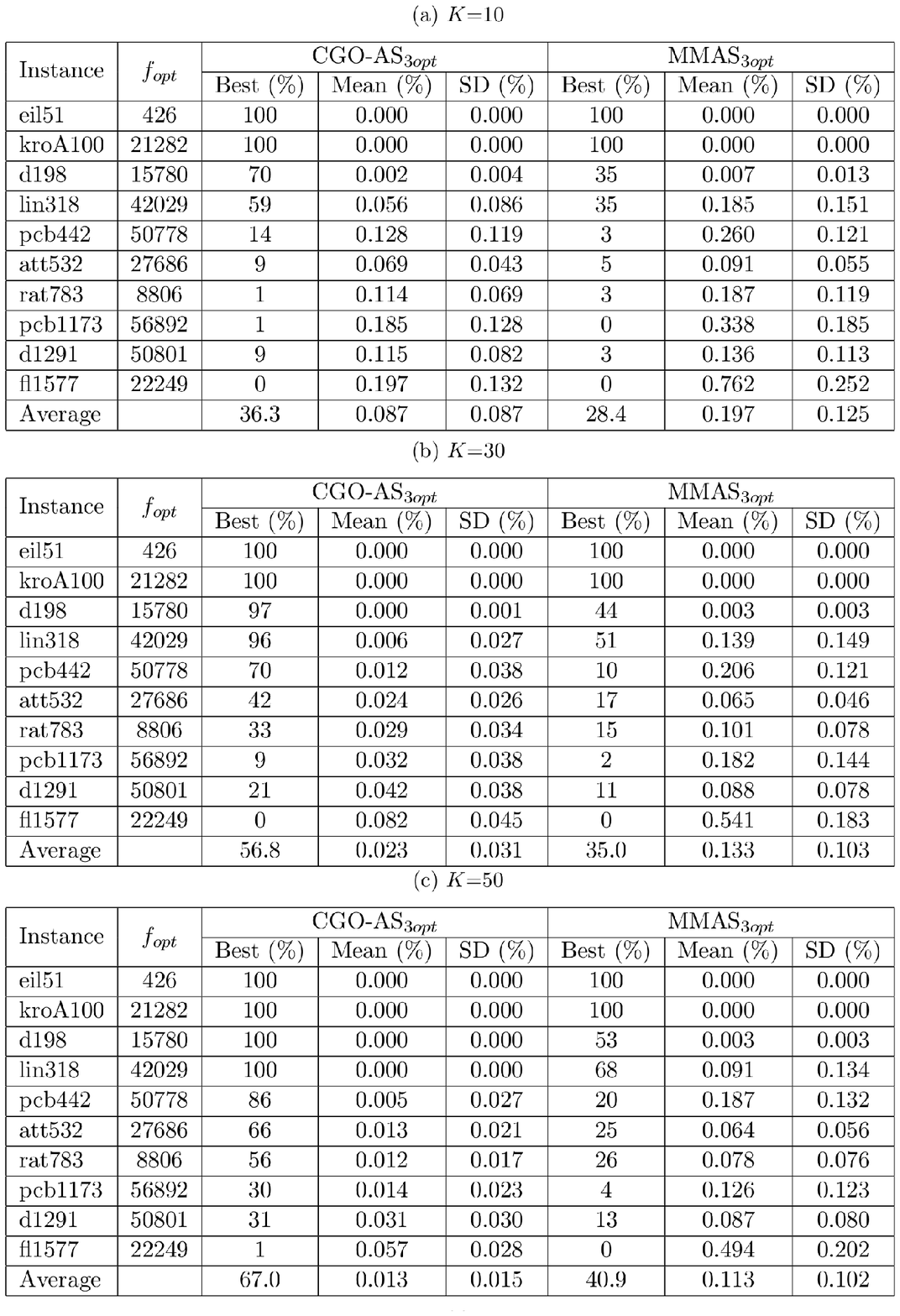} \par}
\end{table}

\end{appendices}

\changespace{1}
%\section*{References}
\bibliographystyle{elsevier}
% \bibliography{bib/own,bib/ants,bib/aco,bib/group,bib/insect,bib/TSP,bib/permutation,bib/cogarc,bib/swarm,bib/de,bib/landscape}

\begin{thebibliography}{}

\bibitem[\protect\citeauthoryear{Anderson}{Anderson}{2005}]{Anderson:2005p1258}
Anderson, J.~R. (2005).
\newblock Human symbol manipulation within an integrated cognitive
  architecture.
\newblock {\em Cognitive Science\/}~{\em 29\/}(3), 313--341.

\bibitem[\protect\citeauthoryear{Applegate, Cook, and Rohe}{Applegate
  et~al.}{2003}]{Applegate:2003p2830}
Applegate, D., W.~Cook, and A.~Rohe (2003).
\newblock Chained Lin-Kernighan for large traveling salesman problems.
\newblock {\em INFORMS Journal on Computing\/}~{\em 15\/}(1), 82--92.

\bibitem[\protect\citeauthoryear{Bentley}{Bentley}{1992}]{bentley1992fast}
Bentley, J.~J. (1992).
\newblock Fast algorithms for geometric traveling salesman problems.
\newblock {\em ORSA Journal on computing\/}~{\em 4\/}(4), 387--411.

\bibitem[\protect\citeauthoryear{Bolek and Wolf}{Bolek and
  Wolf}{2015}]{bolek2015food}
Bolek, S. and H.~Wolf (2015).
\newblock Food searches and guiding structures in {North African} desert ants,
  cataglyphis.
\newblock {\em Journal of Comparative Physiology A\/}~{\em 201\/}(6), 631--644.

\bibitem[\protect\citeauthoryear{Bonabeau, Dorigo, and Theraulaz}{Bonabeau
  et~al.}{2000}]{Bonabeau:2000p3169}
Bonabeau, E., M.~Dorigo, and G.~Theraulaz (2000).
\newblock Inspiration for optimization from social insect behaviour.
\newblock {\em Nature\/}~{\em 406}, 39--42.

\bibitem[\protect\citeauthoryear{Boyd, Richerson, and Henrich}{Boyd
  et~al.}{2011}]{Boyd2011}
Boyd, R., P.~J. Richerson, and J.~Henrich (2011).
\newblock The cultural niche: Why social learning is essential for human
  adaptation.
\newblock {\em Proceedings of the National Academy of Sciences\/}~{\em 108},
  10918--10925.

\bibitem[\protect\citeauthoryear{Bullnheimer, Hartl, and Strauss}{Bullnheimer
  et~al.}{1999}]{bullnheimer1999improved}
Bullnheimer, B., R.~F. Hartl, and C.~Strauss (1999).
\newblock An improved ant system algorithm for the vehicle routing problem.
\newblock {\em Annals of Operations Research\/}~{\em 89}, 319--328.

\bibitem[\protect\citeauthoryear{Cheng, Schultheiss, Schwarz, Wystrach, and
  Wehner}{Cheng et~al.}{2014}]{cheng2014beginnings}
Cheng, K., P.~Schultheiss, S.~Schwarz, A.~Wystrach, and R.~Wehner (2014).
\newblock Beginnings of a synthetic approach to desert ant navigation.
\newblock {\em Behavioural Processes\/}~{\em 102}, 51--61.

\bibitem[\protect\citeauthoryear{Collett}{Collett}{2012}]{collett2012navigational}
Collett, M. (2012).
\newblock How navigational guidance systems are combined in a desert ant.
\newblock {\em Current Biology\/}~{\em 22\/}(10), 927--932.

\bibitem[\protect\citeauthoryear{Collett}{Collett}{2014}]{collett2014desert}
Collett, M. (2014).
\newblock A desert ant's memory of recent visual experience and the control of
  route guidance.
\newblock {\em Proceedings of the Royal Society of London B: Biological
  Sciences\/}~{\em 281\/}(1787), 20140634.

\bibitem[\protect\citeauthoryear{Collett, Collett, Bisch, and Wehner}{Collett
  et~al.}{1998}]{collett1998local}
Collett, M., T.~S. Collett, S.~Bisch, and R.~Wehner (1998).
\newblock Local and global vectors in desert ant navigation.
\newblock {\em Nature\/}~{\em 394\/}(6690), 269--272.

\bibitem[\protect\citeauthoryear{Collett and Collett}{Collett and
  Collett}{2002}]{Collett:2002p3603}
Collett, T. and M.~Collett (2002).
\newblock Memory use in insect visual navigation.
\newblock {\em Nature Reviews Neuroscience\/}~{\em 3\/}(7), 542--552.

\bibitem[\protect\citeauthoryear{Collett, Graham, and Durier}{Collett
  et~al.}{2003}]{Collett:2003p3648}
Collett, T., P.~Graham, and V.~Durier (2003).
\newblock Route learning by insects.
\newblock {\em Current Opinion in Neurobiology\/}~{\em 13\/}(6), 718--725.

\bibitem[\protect\citeauthoryear{Czaczkes, Gr{\"u}ter, Ellis, Wood, and
  Ratnieks}{Czaczkes et~al.}{2013}]{czaczkes2013ant}
Czaczkes, T.~J., C.~Gr{\"u}ter, L.~Ellis, E.~Wood, and F.~L. Ratnieks (2013).
\newblock Ant foraging on complex trails: Route learning and the role of trail
  pheromones in {Lasius} niger.
\newblock {\em The Journal of Experimental Biology\/}~{\em 216\/}(2), 188--197.

\bibitem[\protect\citeauthoryear{Czaczkes, Gr{\"u}ter, and Ratnieks}{Czaczkes
  et~al.}{2015}]{czaczkes2015trail}
Czaczkes, T.~J., C.~Gr{\"u}ter, and F.~L.~W. Ratnieks (2015).
\newblock Trail pheromones: An integrative view of their role in social insect
  colony organization.
\newblock {\em Annual Review of Entomology\/}~{\em 60}, 581--599.

\bibitem[\protect\citeauthoryear{Danchin, Giraldeau, Valone, and
  Wagner}{Danchin et~al.}{2004}]{Danchin:2004p1204}
Danchin, {\'E}., L.-A. Giraldeau, T.~Valone, and R.~Wagner (2004).
\newblock Public information: From nosy neighbors to cultural evolution.
\newblock {\em Science\/}~{\em 305\/}(5683), 487--491.

\bibitem[\protect\citeauthoryear{Deneubourg, Aron, Goss, and
  Pasteels}{Deneubourg et~al.}{1990}]{Deneubourg:1990p3701}
Deneubourg, J., S.~Aron, S.~Goss, and J.~Pasteels (1990).
\newblock The self-organizing exploratory pattern of the {Argentine} ant.
\newblock {\em Journal of Insect Behavior\/}~{\em 3\/}(2), 159--168.

\bibitem[\protect\citeauthoryear{Dennis and Valacich}{Dennis and
  Valacich}{1993}]{Dennis:1993p1298}
Dennis, A. and J.~Valacich (1993).
\newblock Computer brainstorms: more heads are better than one.
\newblock {\em Journal of Applied Psychology\/}~{\em 78\/}(4), 531--537.

\bibitem[\protect\citeauthoryear{Dorigo and Gambardella}{Dorigo and
  Gambardella}{1997}]{Dorigo:1997p2698}
Dorigo, M. and L.~Gambardella (1997).
\newblock Ant colony system: A cooperative learning approach to the traveling
  salesman problem.
\newblock {\em IEEE Transactions on Evolutionary Computation\/}~{\em 1\/}(1),
  53--66.

\bibitem[\protect\citeauthoryear{Dorigo, Maniezzo, and Colorni}{Dorigo
  et~al.}{1996}]{Dorigo:1996p3275}
Dorigo, M., V.~Maniezzo, and A.~Colorni (1996).
\newblock Ant system: Optimization by a colony of cooperating agents.
\newblock {\em IEEE Transactions on Systems Man and Cybernetics Part
  B-Cybernetics\/}~{\em 26\/}(1), 29--41.

\bibitem[\protect\citeauthoryear{Ericsson and Kintsch}{Ericsson and
  Kintsch}{1995}]{Ericsson:1995p1364}
Ericsson, K.~A. and W.~Kintsch (1995).
\newblock Long-term working memory.
\newblock {\em Psychological Review\/}~{\em 102\/}(2), 211--245.

\bibitem[\protect\citeauthoryear{Galef}{Galef}{1995}]{Galef:1995p1128}
Galef, B.~G. (1995).
\newblock Why behaviour patterns that animals learn socially are locally
  adaptive.
\newblock {\em Animal Behaviour\/}~{\em 49\/}(5), 1325--1334.

\bibitem[\protect\citeauthoryear{Giurfa and Capaldi}{Giurfa and
  Capaldi}{1999}]{Giurfa:1999p3614}
Giurfa, M. and E.~Capaldi (1999).
\newblock Vectors, routes and maps: new discoveries about navigation in
  insects.
\newblock {\em Trends in Neurosciences\/}~{\em 22\/}(6), 237--242.

\bibitem[\protect\citeauthoryear{Glenberg}{Glenberg}{1997}]{Glenberg:1997p1390}
Glenberg, A.~M. (1997).
\newblock What memory is for.
\newblock {\em Behavioral and Brain Sciences\/}~{\em 20\/}(1), 1--55.

\bibitem[\protect\citeauthoryear{Goncalo and Staw}{Goncalo and
  Staw}{2006}]{Goncalo:2006p1071}
Goncalo, J.~A. and B.~M. Staw (2006).
\newblock Individualism--collectivism and group creativity.
\newblock {\em Organizational Behavior and Human Decision Processes\/}~{\em
  100}, 96--109.

\bibitem[\protect\citeauthoryear{Goss, Aron, Deneubourg, and Pasteels}{Goss
  et~al.}{1989}]{Goss:1989p3685}
Goss, S., S.~Aron, J.~Deneubourg, and J.~Pasteels (1989).
\newblock Self-organized shortcuts in the {Argentine} ant.
\newblock {\em Naturwissenschaften\/}~{\em 76\/}(12), 579--581.

\bibitem[\protect\citeauthoryear{Gr{\"u}ter, Czaczkes, and Ratnieks}{Gr{\"u}ter
  et~al.}{2011}]{gruter2011decision}
Gr{\"u}ter, C., T.~J. Czaczkes, and F.~L. Ratnieks (2011).
\newblock Decision making in ant foragers (lasius niger) facing conflicting
  private and social information.
\newblock {\em Behavioral Ecology and Sociobiology\/}~{\em 65\/}(2), 141--148.

\bibitem[\protect\citeauthoryear{Gunduz, Kiran, and {\"O}zceylan}{Gunduz
  et~al.}{2015}]{gunduz2015hierarchic}
Gunduz, M., M.~S. Kiran, and E.~{\"O}zceylan (2015).
\newblock A hierarchic approach based on swarm intelligence to solve the
  traveling salesman problem.
\newblock {\em Turkish Journal of Electrical Engineering \& Computer
  Sciences\/}~{\em 23}, 103--117.

\bibitem[\protect\citeauthoryear{Harris, de~Ibarra, Graham, and Collett}{Harris
  et~al.}{2005}]{Harris:2005p3555}
Harris, R., N.~de~Ibarra, P.~Graham, and T.~Collett (2005).
\newblock Ant navigation: Priming of visual route memories.
\newblock {\em Nature\/}~{\em 438\/}(7066), 302--302.

\bibitem[\protect\citeauthoryear{Helsgaun}{Helsgaun}{2000}]{Helsgaun:2000p2779}
Helsgaun, K. (2000).
\newblock An effective implementation of the lin-kernighan traveling salesman
  heuristic.
\newblock {\em European Journal of Operational Research\/}~{\em 126\/}(1),
  106--130.

\bibitem[\protect\citeauthoryear{H{\"o}lldobler and Wilson}{H{\"o}lldobler and
  Wilson}{1990}]{Holldobler:1990p3697}
H{\"o}lldobler, B. and E.~Wilson (1990).
\newblock {\em The Ants}.
\newblock Harvard University Press.

\bibitem[\protect\citeauthoryear{Jackson and Ratnieks}{Jackson and
  Ratnieks}{2006}]{Jackson:2006p3563}
Jackson, D. and F.~Ratnieks (2006).
\newblock Communication in ants.
\newblock {\em Current Biology\/}~{\em 16\/}(15), R570--R574.

\bibitem[\protect\citeauthoryear{Kauffman and Levin}{Kauffman and
  Levin}{1987}]{kauffman1987towards}
Kauffman, S. and S.~Levin (1987).
\newblock Towards a general theory of adaptive walks on rugged landscapes.
\newblock {\em Journal of Theoretical Biology\/}~{\em 128\/}(1), 11--45.

\bibitem[\protect\citeauthoryear{Kavadias and Sommer}{Kavadias and
  Sommer}{2009}]{kavadias2009effects}
Kavadias, S. and S.~C. Sommer (2009).
\newblock The effects of problem structure and team diversity on brainstorming
  effectiveness.
\newblock {\em Management Science\/}~{\em 55\/}(12), 1899--1913.

\bibitem[\protect\citeauthoryear{Liao, Socha, Montes~de Oca, Stutzle, Dorigo,
  et~al.}{Liao et~al.}{2014}]{liao2014ant}
Liao, T., K.~Socha, M.~Montes~de Oca, T.~Stutzle, M.~Dorigo, et~al. (2014).
\newblock Ant colony optimization for mixed-variable optimization problems.
\newblock {\em IEEE Transactions on Evolutionary Computation\/}~{\em 18\/}(4),
  503--518.

\bibitem[\protect\citeauthoryear{Lin and Kernighan}{Lin and
  Kernighan}{1973}]{lin1973effective}
Lin, S. and B.~W. Kernighan (1973).
\newblock An effective heuristic algorithm for the traveling-salesman problem.
\newblock {\em Operations research\/}~{\em 21\/}(2), 498--516.

\bibitem[\protect\citeauthoryear{Macquart, Garnier, Combe, and
  Beugnon}{Macquart et~al.}{2006}]{Macquart:2006p3720}
Macquart, D., L.~Garnier, M.~Combe, and G.~Beugnon (2006).
\newblock Ant navigation en route to the goal: signature routes facilitate
  way-finding of gigantiops destructor.
\newblock {\em Journal of Comparative Physiology A\/}~{\em 192\/}(3), 221--234.

\bibitem[\protect\citeauthoryear{Mahi, Baykan, and Kodaz}{Mahi
  et~al.}{2015}]{mahi2015new}
Mahi, M., {\"O}.~K. Baykan, and H.~Kodaz (2015).
\newblock A new hybrid method based on particle swarm optimization, ant colony
  optimization and 3-opt algorithms for traveling salesman problem.
\newblock {\em Applied Soft Computing\/}~{\em 30}, 484--490.

\bibitem[\protect\citeauthoryear{Merz and Freisleben}{Merz and
  Freisleben}{2001}]{Merz:2001p2780}
Merz, P. and B.~Freisleben (2001).
\newblock Memetic algorithms for the traveling salesman problem.
\newblock {\em Complex Systems\/}~{\em 13\/}(4), 297--345.

\bibitem[\protect\citeauthoryear{Morgan}{Morgan}{2009}]{Morgan:2009p3644}
Morgan, E. (2009).
\newblock Trail pheromones of ants.
\newblock {\em Physiological Entomology\/}~{\em 34\/}(1), 1--17.

\bibitem[\protect\citeauthoryear{Nallaperuma, Wagner, and Neumann}{Nallaperuma
  et~al.}{2015}]{nallaperuma2015analyzing}
Nallaperuma, S., M.~Wagner, and F.~Neumann (2015).
\newblock Analyzing the effects of instance features and algorithm parameters
  for {Max-Min} ant system and the traveling salesperson problem.
\newblock {\em Frontiers in Robotics and AI\/}~{\em 2}, 18.

\bibitem[\protect\citeauthoryear{Nemeth}{Nemeth}{1986}]{Nemeth:1986p980}
Nemeth, C.~J. (1986).
\newblock Differential contributions of majority and minority influence.
\newblock {\em Psychological Review\/}~{\em 93\/}(1), 23--32.

\bibitem[\protect\citeauthoryear{Ouaarab, Ahiod, and Yang}{Ouaarab
  et~al.}{2014}]{ouaarab2014discrete}
Ouaarab, A., B.~Ahiod, and X.-S. Yang (2014).
\newblock Discrete cuckoo search algorithm for the travelling salesman problem.
\newblock {\em Neural Computing and Applications\/}~{\em 24\/}(7-8),
  1659--1669.

\bibitem[\protect\citeauthoryear{Paulus}{Paulus}{2000}]{Paulus:2000p1114}
Paulus, P.~B. (2000).
\newblock Groups, teams, and creativity: the creative potential of
  idea-generating groups.
\newblock {\em Applied Psychology: an International Review\/}~{\em 49\/}(2),
  237--262.

\bibitem[\protect\citeauthoryear{Reinelt}{Reinelt}{1991}]{Reinelt:1991p2617}
Reinelt, G. (1991).
\newblock {TSPLIB} - a traveling salesman problem library.
\newblock {\em ORSA Journal on Computing\/}~{\em 3}, 376--384.

\bibitem[\protect\citeauthoryear{Saenphon, Phimoltares, and Lursinsap}{Saenphon
  et~al.}{2014}]{saenphon2014combining}
Saenphon, T., S.~Phimoltares, and C.~Lursinsap (2014).
\newblock Combining new fast opposite gradient search with ant colony
  optimization for solving travelling salesman problem.
\newblock {\em Engineering Applications of Artificial Intelligence\/}~{\em 35},
  324--334.

\bibitem[\protect\citeauthoryear{Sommer, von Beeren, and Wehner}{Sommer
  et~al.}{2008}]{sommer2008multiroute}
Sommer, S., C.~von Beeren, and R.~Wehner (2008).
\newblock Multiroute memories in desert ants.
\newblock {\em Proceedings of the National Academy of Sciences\/}~{\em
  105\/}(1), 317--322.

\bibitem[\protect\citeauthoryear{St{\"u}tzle and Hoos}{St{\"u}tzle and
  Hoos}{2000}]{Stutzle:2000p2978}
St{\"u}tzle, T. and H.~Hoos (2000).
\newblock {MAX-MIN} ant system.
\newblock {\em Future Generation Computer Systems\/}~{\em 16\/}(8), 889--914.

\bibitem[\protect\citeauthoryear{Tomasello, Kruger, and Ratner}{Tomasello
  et~al.}{1993}]{Tomasello:1993p1330}
Tomasello, M., A.~Kruger, and H.~Ratner (1993).
\newblock Cultural learning.
\newblock {\em Behavioral and Brain Sciences\/}~{\em 16\/}(3), 495--511.

\bibitem[\protect\citeauthoryear{Wehner}{Wehner}{2003}]{Wehner:2003p3574}
Wehner, R. (2003).
\newblock Desert ant navigation: How miniature brains solve complex tasks.
\newblock {\em Journal of Comparative Physiology A\/}~{\em 189\/}(8), 579--588.

\bibitem[\protect\citeauthoryear{Wehner, Boyer, Loertscher, Sommer, and
  Menzi}{Wehner et~al.}{2006}]{Wehner:2006p3740}
Wehner, R., M.~Boyer, F.~Loertscher, S.~Sommer, and U.~Menzi (2006).
\newblock Ant navigation: One-way routes rather than maps.
\newblock {\em Current Biology\/}~{\em 16\/}(1), 75--79.

\bibitem[\protect\citeauthoryear{Wolf}{Wolf}{2008}]{Wolf:2008p3750}
Wolf, H. (2008).
\newblock Desert ants adjust their approach to a foraging site according to
  experience.
\newblock {\em Behavioral Ecology and Sociobiology\/}~{\em 62\/}(3), 415--425.

\bibitem[\protect\citeauthoryear{Wystrach, Mangan, Philippides, and
  Graham}{Wystrach et~al.}{2013}]{wystrach2013snapshots}
Wystrach, A., M.~Mangan, A.~Philippides, and P.~Graham (2013).
\newblock Snapshots in ants? {New} interpretations of paradigmatic experiments.
\newblock {\em The Journal of Experimental Biology\/}~{\em 216\/}(10),
  1766--1770.

\bibitem[\protect\citeauthoryear{Xie, Liu, and Wang}{Xie
  et~al.}{2014}]{xie2014cooperative}
Xie, X.-F., J.~Liu, and Z.-J. Wang (2014).
\newblock A cooperative group optimization system.
\newblock {\em Soft Computing\/}~{\em 18\/}(3), 469--495.

\bibitem[\protect\citeauthoryear{Zhou, Luo, Chen, He, and Wu}{Zhou
  et~al.}{2015}]{zhou2015discrete}
Zhou, Y., Q.~Luo, H.~Chen, A.~He, and J.~Wu (2015).
\newblock A discrete invasive weed optimization algorithm for solving traveling
  salesman problem.
\newblock {\em Neurocomputing\/}~{\em 151}, 1227--1236.

\end{thebibliography}

\end{document}